\documentclass{article}

\usepackage[preprint]{neurips_2023}

\usepackage[utf8]{inputenc} % allow utf-8 input
\usepackage[T1]{fontenc}    % use 8-bit T1 fonts
\usepackage{hyperref}       % hyperlinks
\usepackage{url}            % simple URL typesetting
\usepackage{booktabs}       % professional-quality tables
\usepackage{amsfonts}       % blackboard math symbols
\usepackage{nicefrac}       % compact symbols for 1/2, etc.
\usepackage{microtype}      % microtypography
\usepackage{graphicx}
\usepackage{subcaption}
\usepackage{apalike}
\usepackage{chngpage}
\usepackage{multirow}

\usepackage{amsmath}
\usepackage{amssymb}
\usepackage{mathtools}
\usepackage{amsthm}
\usepackage{makecell}

\usepackage[capitalize,noabbrev]{cleveref}

\newcommand{\rn}[1]{\expandafter{\romannumeral #1\relax}}

\newcommand{\spactor}{\textsc{SpacTor}}

\title{{\spactor-T5}: Pre-training T5 Models with Span Corruption and Replaced Token Detection}

\author{%
  Ke Ye \\
  Google Research\\
  \texttt{kkye@google.com} \\
  \And
  Heinrich Jiang \\
  Google Research\\
  \texttt{heinrichj@google.com} \\
  \And
  Afshin Rostamizadeh \\
  Google Research\\
  \texttt{rostami@google.com} \\
  \And
  Ayan Chakrabarti \\
  Google Research\\
  \texttt{ayanchakrab@google.com} \\
  \And
  Giulia DeSalvo \\
  Google Research\\
  \texttt{giuliad@google.com} \\
  \And
  Jean-Fran\c{c}ois Kagy \\
  Google Research\\
  \texttt{jfkagy@google.com} \\
  \And
  Lazaros Karydas \\
  Google Research\\
  \texttt{lkary@google.com} \\
  \And
  Gui Citovsky \\
  Google Research\\
  \texttt{gcitovsky@google.com} \\
  \And
  Sanjiv Kumar \\
  Google Research\\
  \texttt{sanjivk@google.com} \\
}

\begin{document}

\maketitle

\begin{abstract}
Pre-training large language models is known to be extremely resource intensive and oftentimes inefficient, under-utilizing the information encapsulated in the training text sequences. In this paper, we present \spactor, a new training procedure consisting of (1) a hybrid objective combining span corruption (SC) and token replacement detection (RTD), and (2) a two-stage curriculum that  optimizes the hybrid objective over the initial $\tau$ iterations, then transitions to standard SC loss. We show empirically that the effectiveness of the hybrid objective is tied to the two-stage pre-training schedule, and provide extensive analysis on why this is the case.  In our experiments with encoder-decoder architectures (T5) on a variety of NLP tasks, \spactor-T5 yields the same downstream performance as standard SC pre-training, while enabling a 50\% reduction in pre-training iterations and 40\% reduction in total FLOPs. Alternatively, given the same amount of computing budget, we find that \spactor\ results in significantly improved downstream benchmark performance.
\end{abstract}

\section{Introduction}
\label{sec:introduction}

The recent emergence of successful large language models (LLMs) is in no small part due to the remarkable effectiveness of self-supervised pre-training on massive text corpora. Pre-trained models are found to perform strongly on a wide range of downstream tasks including natural language understanding (NLU) and generation (NLG) --- through fine-tuning on small task-specific datasets~\citep{wei2021finetuned, sanh2022multitask,chung2022scaling}, or through zero-shot / few-shot evaluation,  whereby the model is given only task-specific instructions as input, or a handful of additional exemplars to learn from, respectively~\citep{brown2020language}.

On the one hand, pre-training LLMs using self-supervised objectives frees us from the burden of gathering human labels; on the other, the indirect nature of the supervision also means that each batch of text provides only weak signals that the model can learn from. Consequently, LLMs need to be pre-trained on datasets several orders of magnitude larger than the labeled domain specific datasets. Therefore, a major bottleneck in developing performant LLMs is the massive computational cost incurred at the pre-training phase --- \textit{e.g.}, GPT-3 (175B parameters)~\citep{brown2020language} and PaLM (540B parameters)~\citep{Palm-2022} need up to tens of thousands of PetaFLOP/s-days of compute for pre-training, respectively. In order to effectively scale language models towards better quality, it is imperative to design more efficient self-supervision strategies under which more useful signals for learning downstream tasks are extracted out of each pre-training iteration on unlabeled data

\begin{figure}
    \centering
    \includegraphics[width=0.9\linewidth]{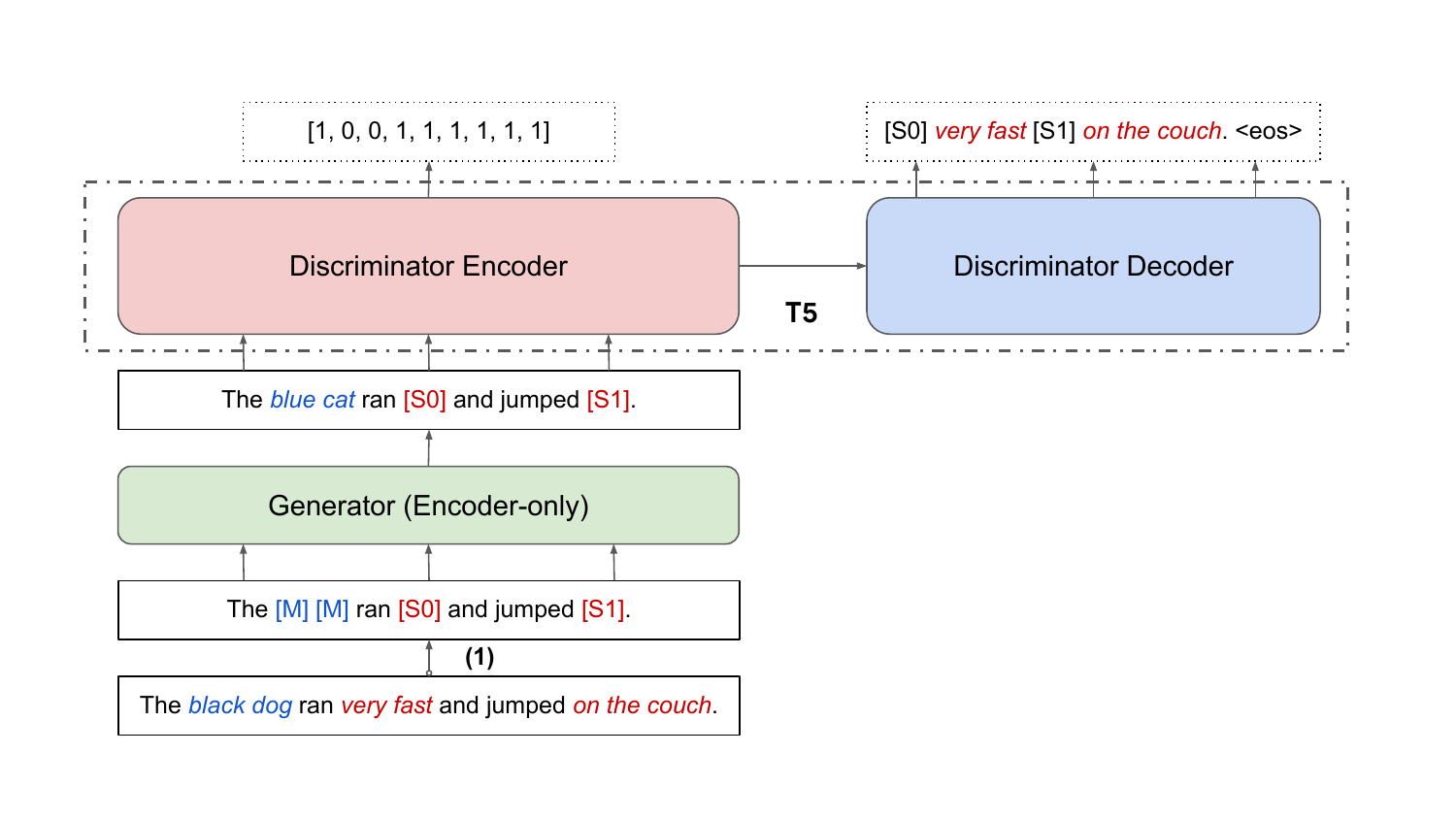}
    \caption{The \spactor\ pre-training objective in the first stage. In step (1), the original text is randomly corrupted with span corruption (marked as {\tt [S0]}, {\tt [S1]}, \emph{etc}, ) and then token-level random masking (marked as {\tt [M]}). A small auxiliary generator model $G$ is trained to recover {\tt [M]} only. The resulting text is then fed into the T5 discriminator $D$, whose encoder component learns to predict at every position whether the token is a replaced one, while its decoder component learns to fill in the ground truth token as in standard span corruption.}
    \label{fig:setup}
\end{figure}

In this paper, we propose \spactor\ (short for ``{\bf Spa}n {\bf c}orruption and {\bf To}ken {\bf r}eplacement"), a new pre-training procedure that significantly improves the efficiency \emph{and} generalization of T5 models~\citep{T5}. \spactor\ consists of two ingredients. The first is an augmentation of the span corruption (SC) pre-training task with the replaced token detection (RTD) objective proposed in ELECTRA \citep{clark2020electra}. The second is a two-staged pre-training schedule: after $\tau$ training steps on hybrid objectives, we continue pre-training only using the vanilla SC objective. The dual task in the first stage is illustrated in Figure~\ref{fig:setup}. Specifically, starting with a span-corrupted input text, an auxiliary generator $G$ replaces a portion of the \emph{uncorrupted} tokens with plausible tokens. The main T5 model (referred to as the discriminator $D$) is pre-trained to detect replaced tokens with its encoder component. Simultaneously, using the same token-replaced input, the discriminator attempts to denoise the SC masks with its decoder. 

\begin{figure*}[t]
\centering  
\begin{subfigure}[b]{0.325\linewidth}
    \centering
    \includegraphics[width=\textwidth]{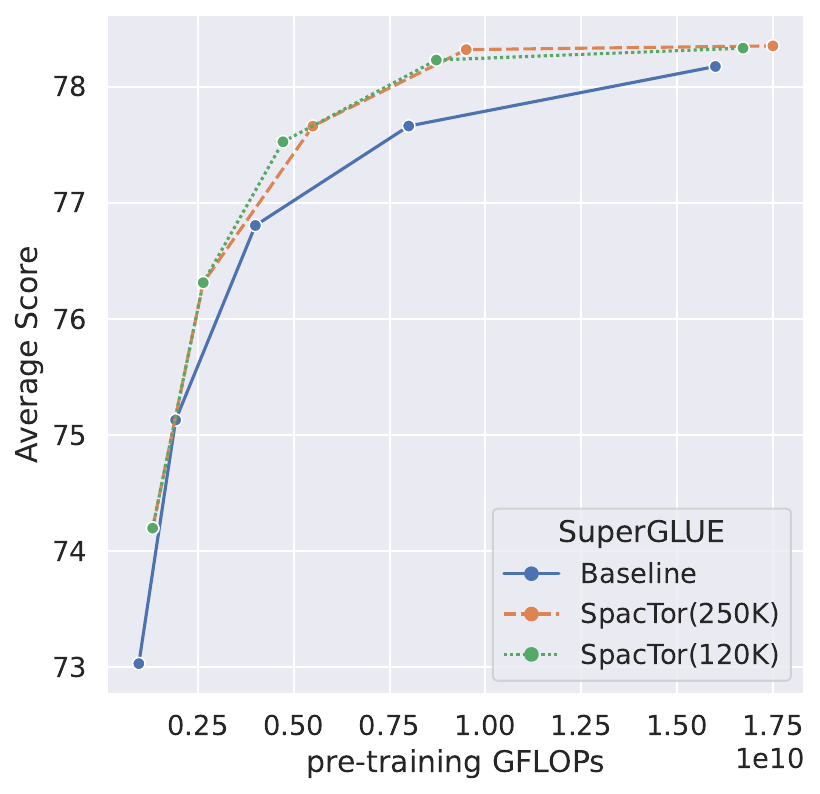}
    \caption{SuperGLUE}
    \label{fig:sglue-flops}
\end{subfigure}
\begin{subfigure}[b]{0.325\linewidth}
    \centering
    \includegraphics[width=\textwidth]{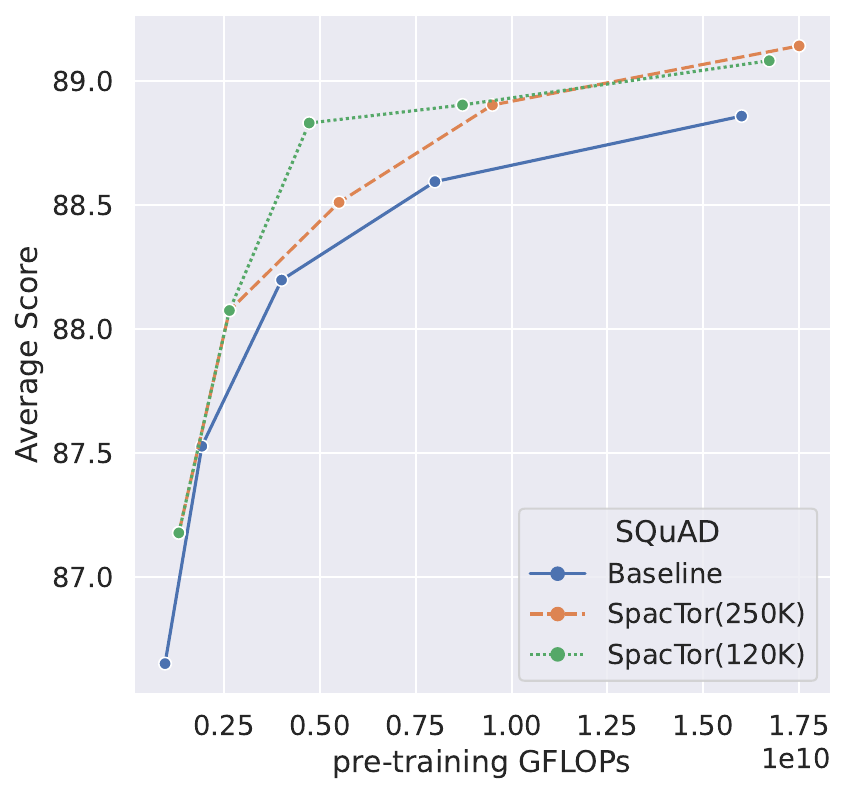}
    \caption{SQuAD}
    \label{fig:squad-flops}
\end{subfigure}
\begin{subfigure}[b]{0.325\linewidth}
    \centering
    \includegraphics[width=\textwidth]{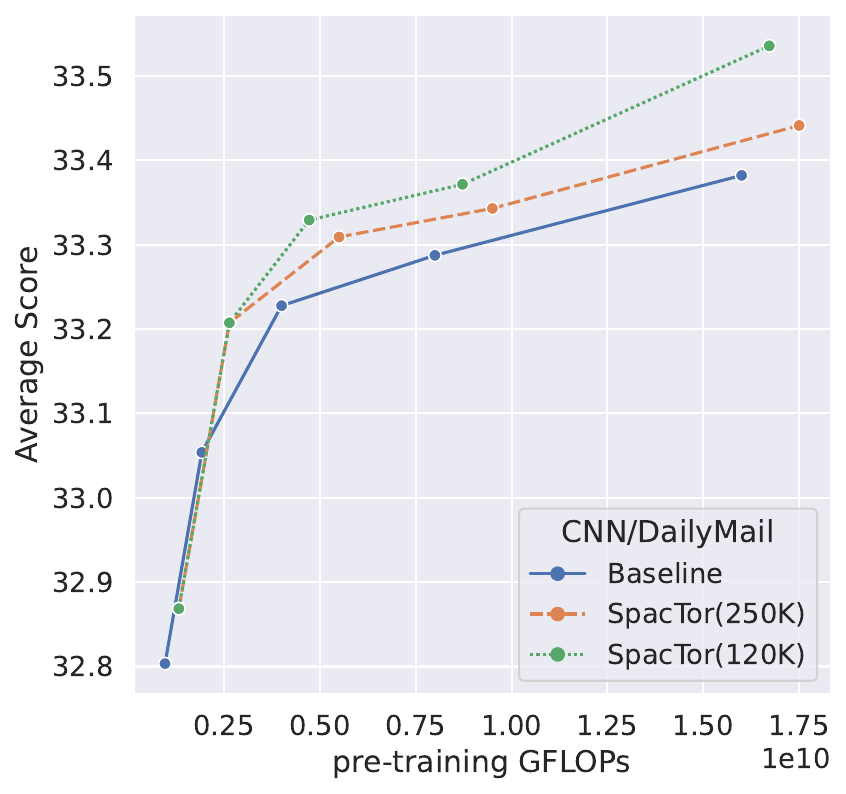}
    \caption{CNN/DailyMail}
    \label{fig:cnndm-flops}
\end{subfigure}
\caption{\spactor($\tau$) performances on SuperGLUE, SQuAD and CNN/DailyMail with respect to pre-training FLOPs. Here, we include \spactor(250K) and \spactor(120K) where the second pre-training stage (using the span corruption objective only) starts at 250K and 120K training steps respectively. The plots for the remaining tasks are presented in Appendix~\ref{app:score-flops-plot}.}
\label{fig:spactor-flops}
\end{figure*}

From a quality standpoint, detecting replaced tokens enforces \emph{all token attention} \citep{clark2020electra}, leading to a better text representation. However, the generator $G$ can also inadvertently introduce misleading yet plausible context (albeit trained non-adversarially), resulting in a noisier training environment for discriminator decoder $D$.\footnote{For example, if we have a corrupted sentence \emph{"Last week we travelled to {\tt [M]}, the capital of {\tt [S0]}."}, where {\tt [M]} is \emph{Tokyo} and {\tt [S0]} is \emph{Japan}. The generator $G$ can reasonably produce a different city for the mask {\tt [M]}, which consequently leads the discriminator to associate it with the capital of Japan due to the use of teacher forcing during training.} As we explain in more detail in Section~\ref{sec:experiments}, the advantages of RTD are predominantly observed in the initial stages of pre-training. As the training progresses however, these benefits are eventually overshadowed by the noise introduced to the discriminator's encoder. This phenomenon naturally motivates the two-staged training, which significantly boosts the performance on various downstream tasks. Figure~\ref{fig:spactor-flops} shows examples of these improvements when $\tau$ equals 120K (1/8 of total iterations) and 250K (1/4 of total iterations) on the SuperGLUE \citep{wang2019superglue}, SQuAD \citep{rajpurkar2016squad} and CNN/DailyMail \citep{hermann2015teaching} benchmarks. These and several other results are discussed in detail in Section~\ref{sec:experiments} and Appendix~\ref{app:score-flops-plot}.

From the perspective of efficiency, one major benefit of our design is that we do not increase the target length of the decoder. A naive extension of the ELECTRA approach to the encoder-decoder architecture would require decoding the entire original input sequence rather than only the corrupted spans, which is undesirable since the complexity of the decoder's self-attention is $\mathcal{O}(L^2)$ for a given target length $L$. The additional computational overhead of \spactor, on the contrary, mainly comes from the inference and back-propagation of the generator $G$ (typically much smaller compared to the discriminator $D$) and a light-weight binary classification head. The cost is only incurred during the first $\tau$ training steps and gets amortized over the rest of the steps. Consequently, \spactor\ achieves a $50\%$ reduction in training iterations and a 40\% reduction in FLOPs while maintaining task performance, as presented in detail in Section~\ref{sec:experiments}. 

The main contribution of the papers are:
\begin{enumerate}
    \item We propose a novel combination of RTD and SC, thus extending ELECTRA to encoder-decoder architecture. 
    \item We analyze extensively the interactions between the two objectives, and establish a two-stage pre-training schedule.
    \item We show that \spactor\ scales well as model size increases, and offers around 40\% savings in total pre-training compute.
\end{enumerate}

\section{\spactor~Method}
\label{sec:method}

In this section, we first describe in detail the pre-training objective of \spactor\ highlighted in Figure \ref{fig:setup}; after that we describe the methodology of two-stage pre-training.

\subsection{The Hybrid Pre-training Objective}

Given an input text composed of a sequence of tokens $X = \{x_0, x_1, ..., x_{N-1}\}$, we introduce two types of masks and apply them sequentially:

\textbf{SC masks} \citep{T5}. Let $X_{i, j}$ be the set of consecutive tokens $X_{i, j} = \{x_i, x_{i+1}, ..., x_{j-1}, x_{j}\}$. SC selects $p$ disjoint spans $\mathcal{S}_p = \{X_{i_k, j_k}\}_{k=0}^{p-1}$ uniformly at random, with average span length $\mu = 3$. Each $X_{i_k, j_k}$ is then replaced with a single sentinel token \texttt{[S$k$]}: 
\begin{equation}
  \begin{split}
    \left\{ x_0, ..., X_{i_0, j_0}, ..., X_{i_k, j_k}, ..., x_{N-1} \right\} \longrightarrow \\[1ex]
    \left\{ x_0,..., \mathtt{[S0]}, ..., \mathtt{[S}k\mathtt{]}, ..., x_{N-1} \right\}.
  \end{split}
\label{eqs:span-corruption}
\end{equation}

For convenience, we denote $X_{\mathrm{c}}$ to be the right hand side of Equation~\ref{eqs:span-corruption}.

\textbf{MLM masks}. For the rest of the tokens $X_{\mathrm{c}}\ \backslash\ \{ \mathtt{[S}k\mathtt{]} \}$, we continue \emph{token level} masking by selecting $q$ additional tokens $\mathcal{M}_q = \{x_{u_m}\}_{m=0}^{q-1}$ uniformly at random and replace them with mask \texttt{[M]}: 
\begin{equation}
  \begin{split}
    \left\{ x_0, ..., x_{u_0},..., \mathtt{[S}k\mathtt{]}, ..., x_{u_m},..., x_{N-1} \right\} \longrightarrow \\[1ex]
    \left\{ x_0, ..., \mathtt{[M]},..., \mathtt{[S}k\mathtt{]}, ..., \mathtt{[M]},..., x_{N-1} \right\}.
  \end{split}
\end{equation}

We denote the final corrupted sentence with both masks as $X_{\mathrm{c}}^{\mathrm{MLM}}$.

Note that we apply MLM masks \emph{after} SC, to utilize well-established SC algorithm and distributions. MLM masks, being at token level, can also be inserted avoiding SC masks naturally.

The inputs are now passed to a generator $G$ and a discriminator $D$. $G$ and $D$ share the same token embedder \citep{clark2020electra} and are jointly trained. 

\textbf{Generator $G$}. The backbone of $G$ is a bidirectional transformer \emph{encoder}, mapping each token in $X_{\mathrm{c}}^{\mathrm{MLM}}$ to contextualized vector representations $\mathbf{H}^G_{d \times n} = \{h^G_0, h^G_1, ..., h^G_{n-1}\}$ where $h^G_{\ell}, \ell = 0,...,n-1$ is a $d$-dimensional column vector and $n = N - p(\mu - 1)$ is the length of $X_{\mathrm{c}}^{\mathrm{MLM}}$. We add a linear projection layer $\mathbf{W}^G_{v\times d}$ that mapping $h^G_{\ell}$ to the $v$-dimensional embedding space of vocabulary. Finally, a softmax is taken to calculate the probabilities of output tokens:
\begin{equation}
    p_G\left(x_{\ell} |\ X_{\mathrm{c}}^{\mathrm{MLM}} \right) = \mathrm{softmax} \left( \mathbf{W} \cdot h^G_{\ell} \right),
\label{eqs:generator-prob}
\end{equation}

The loss function for $G$ is
\begin{equation}
    \mathcal{L}_G = \mathbb{E} \left( \sum_{\ell} - \log p_G\left(x_{\ell} |\ X_{\mathrm{c}}^{\mathrm{MLM}} \right) \right)
\end{equation}

\textbf{Discriminator $D$}. $D$ is a T5 model. The encoder input of $D$ is generated by sampling from categorical distribution $p_G$ and replacing each \texttt{[M]} in $X_{\mathrm{c}}^{\mathrm{MLM}}$ with plausible token $\widehat{x}$. We refer to the resulting text as $\widehat{X_{\mathrm{c}}}$, which is used as the encoder input of $D$.

The encoder output of $D$', $\mathbf{H}^D_{d \times n} = \{h^D_0, h^D_1, ..., h^D_{n-1}\}$, is fed into an MLP layer $f$ followed by sigmoid to determine whether the given token is the same as the ground truth or is replaced:
\begin{equation}
    p_D^{\mathrm{RTD}} (\widehat{x}_{\ell}) = \exp(f(h^D_{\ell})) / \left[ 1+ \exp(f(h^D_{\ell})) \right].
\end{equation}

The corresponding loss for RTD is
\begin{equation}
\mathcal{L}_D^{\mathrm{RTD}} = \mathbb{E} \left[ \sum_{\ell=0}^{n-1} -\mathbb{I}(\widehat{x}_{\ell} = x_{\ell})\log p_D^{\mathrm{RTD}} (\widehat{x}_{\ell}) - \mathbb{I}(\widehat{x}_{\ell} \neq x_{\ell})\log (1-p_D^{\mathrm{RTD}}(\widehat{x}_{\ell}))  \right]
\end{equation}

On the other hand, the decoder of $D$ is trained to find the actual tokens behind the SC masks $\mathtt{[S}k\mathtt{]}$, taking into account the embedding $\mathbf{H}^D_{d \times n}$. As in \citet{T5}, we formulate the decoder target as the concatenation of SC masks and the ground truth tokens:
\begin{equation}
    T := \mathtt{[S}0\mathtt{]}\ X_{i_0, j_0}\ ...\ \mathtt{[S}(p-1)\mathtt{]}\ X_{i_{p-1}, j_{p-1}}\ \mathtt{[EOS]}.
\end{equation}

This gives the following loss,
\begin{equation}
    \mathcal{L}_D^{\mathrm{SC}} = \mathbb{E} \left[ \sum_{i=1}^{p\mu+p+1} -\log p_D^{\mathrm{SC}}\left( T_i\ |\ T_{i-1},\ ...,\ T_0; \widehat{X_{\mathrm{c}}} \right) \right].
\end{equation}

The final loss of training is the weighted sum of three terms:
\begin{equation}
    \mathcal{L} = \mathcal{L}_G + \lambda_1 \mathcal{L}_D^{\mathrm{RTD}} + \lambda_2 \mathcal{L}_D^{\mathrm{SC}}, \ \ \ \lambda_{1, 2} \geq 0.
    \label{eqs:loss}
\end{equation}

\subsection{Two-staged Pre-training}

As described in Section \ref{sec:introduction} and elaborated in Section \ref{subsubsec:single-stage-pretrain} below, the existence of MLM masks, plus the imperfection of the generator $G$ itself may provide misleading context $\widehat{X_{\mathrm{c}}}$ which obstructs training from SC. We therefore introduce a one-parameter generalization that after training hybrid objective with $\tau$ iterations, only the discriminator $D$ and shared token embedder are retained, and continue the rest of the pre-training with vanilla SC objective.

\section{Experiments}
\label{sec:experiments}

In this section, we begin by describing our experimental setup. To emphasize the stage transition $\tau$ and the discriminator size $M$, we explicitly write $\spactor_{M}(\tau)$ in the remaining of the paper. At two extremes, when $\tau = 0$ (resp. $\tau = \infty$), we train with the SC objective (resp. the hybrid objective) exclusively. We then show that the performance gain of $\spactor_{\textrm{Base}}(\infty)$ is not sustainable (Section \ref{subsubsec:single-stage-pretrain}), and a two-staged pre-training is the natural remedy (Section \ref{subsubsec:continued-pretrain}). With the knowledge gained from the Base model up to Section \ref{subsubsec:eff_analysis}, we extend the experiment to the Large model in Section \ref{subsubsec:t5_large}.

\subsection{Setup}

{\bf Pre-training procedures.} We closely follow the convention of the original T5 paper \citep{T5}, and focus on the T5.1.0 model throughout our experiments. The model is pre-trained on Colossal Clean Crawled Corpus (C4), a massive English-only web extracted text set. We use the SentencePiece tokenizer with 32,000 tokens for preprocessing the raw text corpus, and the Adafactor optimizer \citep{adafactor2018} for model parameter training. Details of the pre-training hyper-parameters and their tuning are discussed in Table~\ref{tab:pretrain_params} of Appendix~\ref{app:pre-training}.

{\bf Fine-tuning procedure.} The weights of the pre-trained discriminator $D$ and the token embedder are used to initialize fine-tuning. In accordance with standard practice, we use a constant learning rate and train over a sufficiently large number of iterations to ensure that the validation metrics have converged. More details of the fine-tuning hyperparameters can be found in Appendix~\ref{app:fine-tuning}.

{\bf Evaluation.} We use the T5.1.0 model pre-trained with span corruption only \citep{T5} as baseline. Table \ref{tab:nlp_tasks} gives a list of representative natural language tasks we evaluate in this paper. For tasks having multiple sub-tasks, we treat them independently, and select the best checkpoint based on the maximal value of the average of the corresponding set of metrics. For FLAN instruction-tuning in particular, we focus on the benchmark comprised of 27 tasks from BIG-Bench (BBH) \citep{srivastava2022beyond} and 57 tasks from Massive Multitask Language Understanding (MMLU) \citep{hendrycks2021measuring} with direct answers. Here we do not include benchmarks with Chain-of-Thought~\citep{wei2022chain} as reasoning is an emergent capability of larger models beyond O(10B) scale. We compare the fine-tuning results without using LM adaptation~\citep{lester-etal-2021-power} to directly reflect quality gains. We also exclude tasks involving multilinguality such as WMT translation (\textit{e.g.,} see \citet{barrault-etal-2020-findings}), because those tasks are more suitable for mT5 models \citep{xue2020mt5}. 

\begin{table*}[ht]
\begin{center}
\begin{tabular}{ccccr}
\toprule
\bf{Task} & {\bf Description} & {\bf No. Datasets} & {\bf Reference} \\ \midrule
GLUE &      General Language Understanding  & 7      & \citet{wang2018glue} \\ 
SuperGLUE &    General Language Understanding  &   8   & \citet{wang2019superglue} \\
  SQuAD   &     QA (context)    &    1   & \citet{rajpurkar2016squad} \\
  CNN/DailyMail &  News Summarization & 1  & \citet{hermann2015teaching} \\
  Rainbow & Commonsense Reasoning & 6 & \citet{lourie2021unicorn} \\
  FLAN & Instruction-tuning & 6 & \citet{chung2022scaling} \\
\bottomrule
\end{tabular}
\end{center}
\caption{List of natural language tasks for fine-tuning.}
\label{tab:nlp_tasks}
\end{table*}

\subsection{Results}

We now present the main experimental results for $\spactor_{\textrm{Base}}(\tau)$ and $\spactor_{\textrm{Large}}(\tau)$. For the former, we compare $\tau = \infty$ and $\tau < \infty$ and emphasize the importance of training stage transition. We also analyze the quantitative gains from both generalizability and efficiency perspective.

\subsubsection{Single stage pre-training}
\label{subsubsec:single-stage-pretrain}

\begin{figure}[t]
\centering  
\begin{subfigure}[b]{0.45\linewidth}
    \centering
    \includegraphics[width=\textwidth]{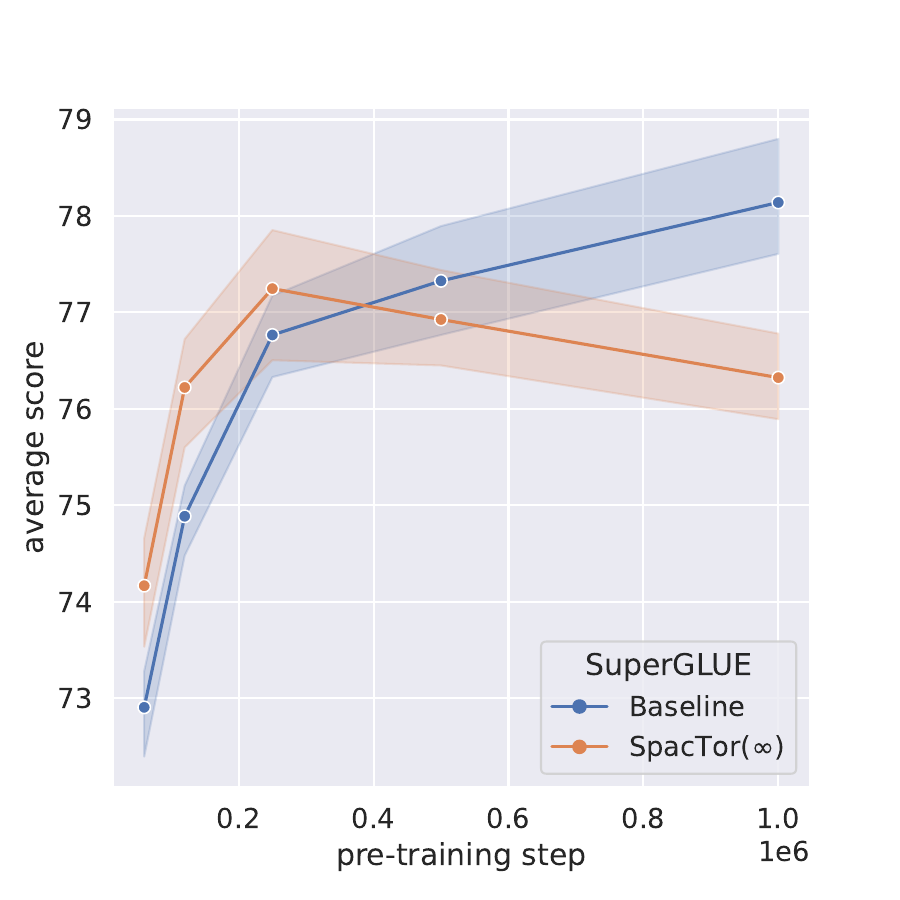}
    \caption{SuperGLUE}
    \label{fig:spactor-single-stage-sglue}
\end{subfigure}\hspace{5mm}%
\begin{subfigure}[b]{0.45\linewidth}
    \centering
    \includegraphics[width=\textwidth]{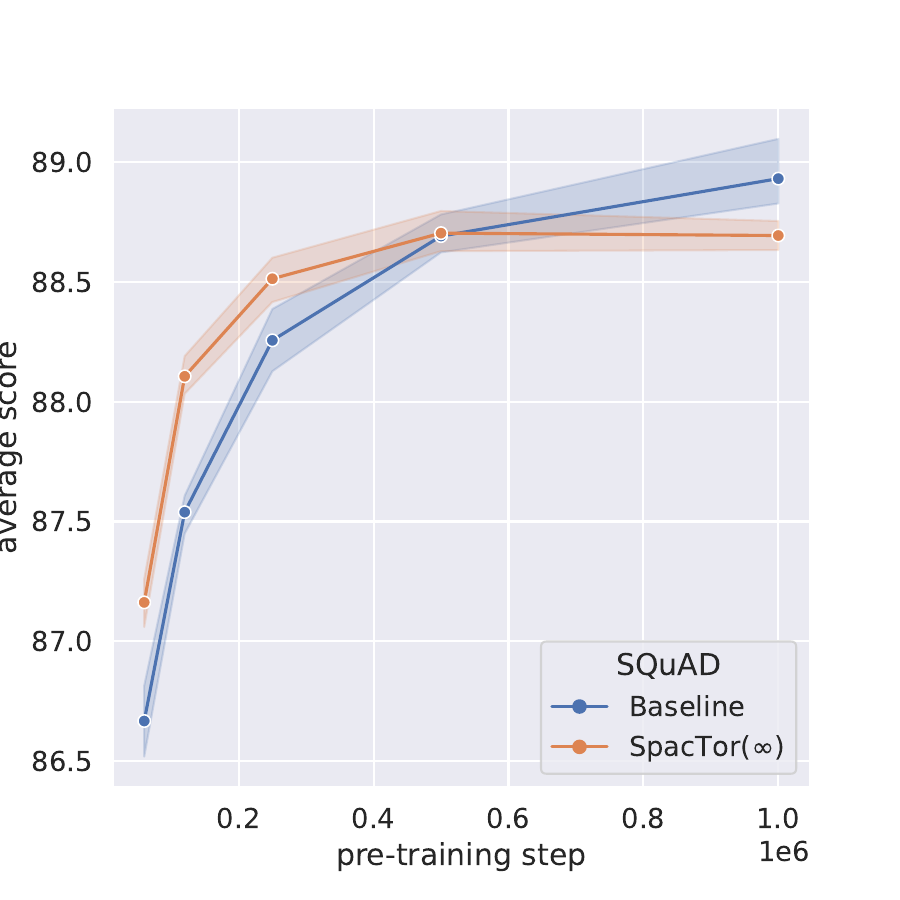}
    \caption{SQuAD}
    \label{fig:spactor-single-stage-squad}
\end{subfigure}
\caption{Average score on downstream tasks ($y$-axis) when continuously fine-tuning along the pre-training checkpoints ($x$-axis). The error band illustrates the min-max range over 5 independent runs.}
\label{fig:spactor-single-stage}
\end{figure}

As motivated in Section \ref{sec:introduction}, jointly pre-training on SC and RTD can be a double-edged sword. This is reflected in Figure~\ref{fig:spactor-single-stage} where we plot the continuous fine-tuning results for $\spactor_{\textrm{Base}}(\infty)$ up to 1M steps. While the added RTD objective enhances performance in the early iterations, the gains vanish after around 250K pre-training steps and the model eventually under-performs compared to the baseline. 

\begin{figure*}[ht]
\centering  
\begin{subfigure}[b]{0.45\linewidth}
    \centering
    \includegraphics[width=\textwidth]{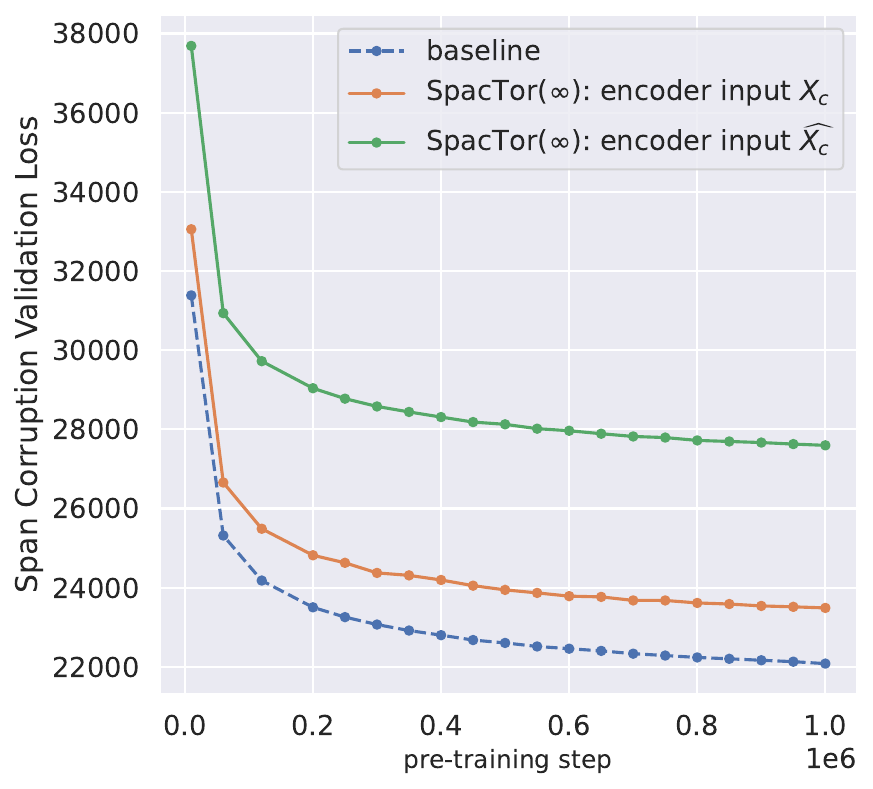}
    \caption{}
    \label{fig:pretraining-val-loss}
\end{subfigure}\hspace{5mm}%
\begin{subfigure}[b]{0.45\linewidth}
    \centering
    \includegraphics[width=\textwidth]{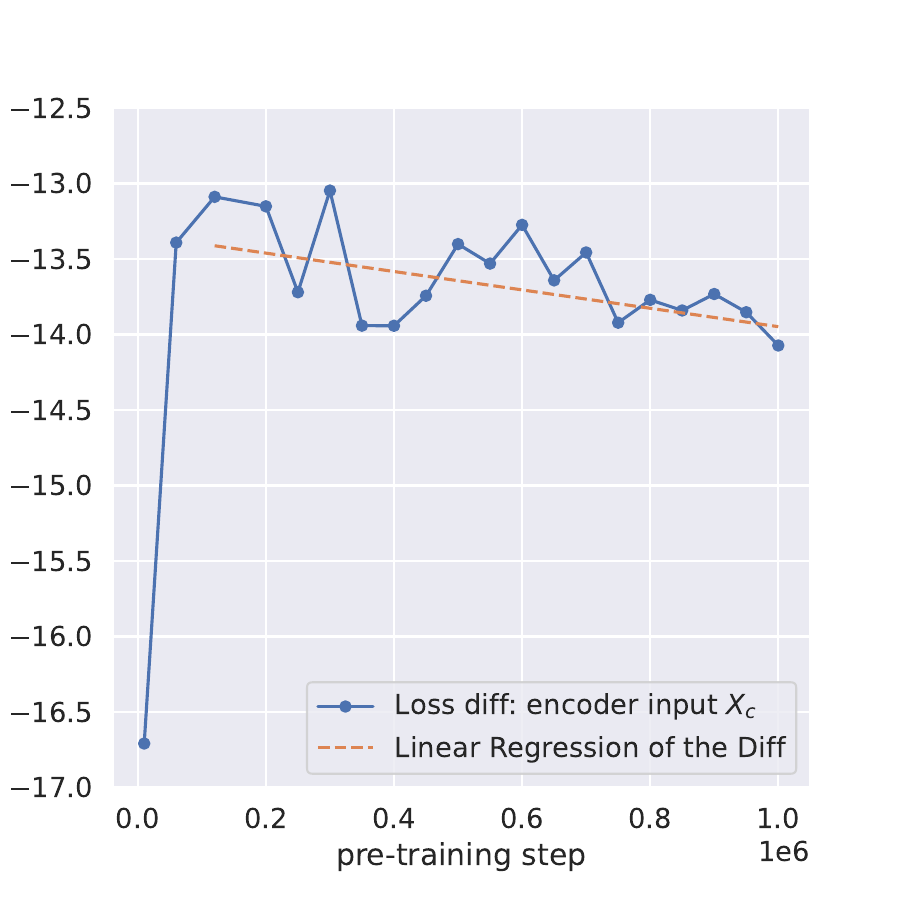}
    \caption{}
    \label{fig:pretraining-val-diff}
\end{subfigure}
    \caption{{\bf (Left)} Validation loss curve for baseline and \spactor($\infty$). {\bf (Right)} Validation cross-entropy loss differences between baseline and \spactor($\infty$) evaluated with encoder input $X_{\mathrm{c}}$. The dashed line is the linear regression fits to the data starting at iteration 120K.}
\label{fig:pretrain-stats}
\end{figure*}

To gain more insights, we compare validation loss $\mathcal{L}_D^{\mathrm{SC}}$ against baseline, when the encoder inputs are the original context $X_{\mathrm{c}}$ or the noisy context $\widehat{X_{\mathrm{c}}}$ respectively in Figure \ref{fig:pretraining-val-loss}. When noisy input $\widehat{X_{\mathrm{c}}}$ is consumed, the loss is noticeably inferior compared to using $X_{\mathrm{c}}$, an indication that replaced tokens in fact hurts the validation score of SC.

In Figure \ref{fig:pretraining-val-diff}, we subtract $\spactor_{\textrm{Base}}(\infty)$'s validation cross entropy against baseline. The gap in loss reduces initially, as the generator $G$ produces more correct tokens. An inflection occurs at around 200K pre-training steps, after that a reverse trend is developed. The trend is statistically significant, based on the hypothesis test carried out in Appendix \ref{app:stats-testing}. This implies the discriminator $D$'s performance on the SC objective is diverging further away from baseline, suggesting that the training is bottlenecked by noise in the input context $\widehat{X_{\mathrm{c}}}$. The inflection point approximately occurs at the same time as the one happened in Figure \ref{fig:spactor-single-stage} --- a qualitative confirmation that downstream metric decay can be attributed to the degradation of span corruption performance during pre-training.

We conjecture that RTD helps in early training iterations because discriminator $D$ is still weak, and correlations of input and target tokens are not yet properly established. Therefore, noise in $G$ does not matter too much. Meanwhile, all token attention enforced by RTD greatly aids the model to maximize the usage of input context, hence boosting the downstream metrics. 

\subsubsection{With continued pre-training}
\label{subsubsec:continued-pretrain}

\begin{table}[ht]
\begin{adjustwidth}{-.7in}{-.7in}
\begin{center}
\footnotesize
\setlength{\tabcolsep}{2.2pt}
\begin{tabular}{cccccccccc}
\toprule
\bf{$\tau$} & \bf{Ckpt.} & {\bf FLOPs} & {\bf GLUE} & {\bf SuperGLUE} & {\bf SQuAD} & {\bf CNNDM} & {\bf Rainbow} & {\bf MMLU} & {\bf BBH} \\ \midrule
0 & 500K & 1.0 & 85.89 $\pm$ 0.17 & 77.33 $\pm$ 0.74 & 88.59 $\pm$ 0.05 & 33.27 $\pm$ 0.12 & 70.14 $\pm$ 0.25 & 50.20 $\pm$ 1.47 & 36.82 $\pm$ 0.55 \\[0.4em]
250K & 500K & 1.2 & {\bf 86.46} $\pm$ 0.17 & {\bf 78.26} $\pm$ 0.63 & {\bf 88.91} $\pm$ 0.11 & {\bf 33.34} $\pm$ 0.10 & {\bf 71.60} $\pm$ 0.18 & {\bf 51.15} $\pm$ 0.80 & {\bf 37.30} $\pm$ 0.33 \\[0.4em]
120K & 500K & 1.1 & {\bf 86.35} $\pm$ 0.13 & {\bf 78.23} $\pm$ 0.81 & {\bf 88.93} $\pm$ 0.11 & {\bf 33.37} $\pm$ 0.10 & {\bf 71.34} $\pm$ 0.23 & {\bf 51.01} $\pm$ 0.40 & {\bf 36.97} $\pm$ 0.31 \\[0.4em]
60K & 500K & 1.05 & {\bf 86.28} $\pm$ 0.22 & {\bf 78.50} $\pm$ 0.56 & {\bf 88.95} $\pm$ 0.14 & 33.27 $\pm$ 0.08 & {\bf 71.35} $\pm$ 0.16 & {\bf 50.67} $\pm$ 1.02 & 36.72 $\pm$ 0.30 \\
\midrule
0 & 1M & 2.0 & 86.11 $\pm$ 0.17 & 78.14 $\pm$ 0.80 & 88.90 $\pm$ 0.23 & 33.34 $\pm$ 0.10 & 71.00 $\pm$ 0.20 & 52.79 $\pm$ 0.95 & 37.57 $\pm$ 0.77 \\[0.4em]
250K & 1M & 2.2 & {\bf 86.48} $\pm$ 0.29 & {\bf 78.33} $\pm$ 0.76 & {\bf 89.09} $\pm$ 0.12 & {\bf 33.47} $\pm$ 0.07 & {\bf 72.27} $\pm$ 0.29 & {\bf 52.96} $\pm$ 0.61 & {\bf 38.18} $\pm$ 0.84 \\[0.4em]
120K & 1M & 2.1 & {\bf 86.57} $\pm$ 0.35 & {\bf 78.16} $\pm$ 0.76 & {\bf 88.99} $\pm$ 0.14 & {\bf 33.53} $\pm$ 0.09 & {\bf 72.14} $\pm$ 0.25 & {\bf 52.81} $\pm$ 0.57 & {\bf 38.08} $\pm$ 0.65 \\
\bottomrule
\end{tabular}
\end{center}
\end{adjustwidth}
\vspace{2mm}
\caption{Average score of each downstream tasks for $\spactor_{\textrm{Base}}(\tau)$. When $\tau=0$ it becomes the baseline. We present both the mean value and standard deviation across five independent runs. We bold numbers for $\spactor_{\textrm{Base}}(\tau)$ with a higher mean than baseline at the same pre-training steps. In the third column, we add the normalized FLOPs where baseline-500K checkpoint is normalized to be 1.0. Details are presented in Section \ref{subsubsec:eff_analysis}. }
\label{tab:finetune_results}
\end{table}

Now we discuss $\tau < \infty$. In practice, based on Figure \ref{fig:spactor-single-stage} and Figure \ref{fig:pretraining-val-diff} we compare cases with $\tau$ to be 60K, 120K or 250K. 

In Table~\ref{tab:finetune_results}, we summarize the downstream task metrics for baseline and $\spactor_{\textrm{Base}}(\tau)$ fine-tuned at 500K / 1M checkpoints. The results show that at 500K checkpoint, $\spactor_{\textrm{Base}}(\tau)$ consistently outperforms the baseline by a large margin. For $\tau = 250$K as an example, the gain is at least one standard deviation, and can reach as large as $3\sigma$ on tasks like GLUE and SQuAD. Except MMLU and BBH, $\spactor_{\textrm{Base}}(\tau)$ with only half of the pre-training iterations achieves similar or even better downstream performances than baseline. When training to 1M, $\spactor_{\textrm{Base}}(\tau)$ retains its superiority over baseline, even though 75\% of the steps are trained with SC only. This implies that the two-staged pre-training, indeed, fixes the decay in performance shown in Figure~\ref{fig:spactor-single-stage}.

Interestingly, comparing the fine-tuning results at the 500K checkpoint when $\tau$ equals 250K, 120K and 60K, we see there is no obvious difference on tasks such as SuperGLUE and SQuAD. For others, reducing $\tau$ from 250K to 60K we see a significant drop in the metrics, some of which become even on par with the baseline. This indicates that 60K iterations is perhaps too early for the transition to the second stage of pre-training. For that reason, we do not evaluate $\spactor_{\textrm{Base}}(60\textrm{K})$ at 1M iterations anymore.

The breakdown of individual subtasks and their evaluation metrics are described in Appendix \ref{app:benchmark-breakdown}.

\subsubsection{Efficiency analysis}
\label{subsubsec:eff_analysis}

Comparing downstream tasks at the same number of iterations (\textit{i.e.} Table~\ref{tab:finetune_results}) is not entirely indicative of training efficiency as $\spactor_{\textrm{Base}}(\tau)$ requires more floating point operations (FLOPs) per step in the first $\tau$ iterations. Nonetheless, as the analysis in this section shows, \spactor\ achieves a net increase in performance as a function of overall compute cost.

We compare the actual compute cost using two approaches. In the first approach, we read sequences per second metric using the T5X library \citep{roberts2022t5x}, a direct reflection of wall clock time. We normalize the value against the baseline to avoid hardware-dependent specifics. In the second approach, we calculate FLOPs per iteration, a hardware independent quantity. As summarized in Table~\ref{tab:flops_count}, we find that pre-training on $\spactor_{\textrm{Base}}(\tau)$ during the first stage incurs about 37.5\% more FLOPs at each iteration than the baseline, which approximately matches the relative value of sequence per second. 

\begin{table}[ht]
\begin{center}
\setlength{\tabcolsep}{15pt}
\begin{tabular}{ccc}
\toprule
\bf{Experiment} & {\bf Seqs / second} & {\bf FLOPs / step} \\ 
\midrule
Baseline & 1.0 & $1.6 \times 10^4$ GFLOPs \\[0.3em]
$\spactor_{\textrm{Base}}(\tau)$ (1st stage) & 0.7 & $2.2 \times 10^4$ GFLOPs \\
\bottomrule
\end{tabular}
\vspace{2mm}
\caption{Efficiency analysis of baseline and $\spactor_{\textrm{Base}}(\tau)$ in the first stage (The second stage compute is identical to the baseline). Seqs / second is normalized using the baseline value.}
\label{tab:flops_count}
\end{center}
\end{table}

In the second column of Table~\ref{tab:finetune_results}, we added the relative FLOPs of each method at a fixed iteration. For example, $\spactor_{\textrm{Base}}(\textrm{250K})$ has an overall normalized FLOPs of $0.5 \times 1.375 + 0.5 \times 1.0 \approx 1.2$ after 500K iterations. For majority of the benchmarks, the 500K checkpoint is matching or beating the baseline 1M ones with a normalized FLOPs of 2.0. This represent an overall efficiency gain of at least 40\%. It is also worth noting that, as the length of the second stage training grows relative to the first stage, the extra cost of $\spactor_{\textrm{Base}}(\tau)$ is reduced. For example, at 1M iterations the number shrinks to $2.2 / 2 = 1.1$.

To better illustrate performance as a function of compute, Figure~\ref{fig:spactor-flops} plots average score of SuperGLUE, SQuAD and CNN/DailyMail with respect to FLOPs. Not only do we see that $\spactor_{\textrm{Base}}(\tau)$ achieves the same average score as baseline-1M with 40\% less compute, but that is also outperforms baseline across the majority of compute budgets. In Appendix \ref{app:score-flops-plot} we include similar plot for the remaining tasks.

\subsubsection{Large models}
\label{subsubsec:t5_large}

We now scale up \spactor\ to T5-Large model \citep{T5} of around 700M parameters. We pick transition parameter $\tau = 120\textrm{K}$ and MLM ratio to be 20\%, due to the proportional size increase of the generator $G$. Other hyperparameters such as coefficients $\lambda_{1,2}$ (Equation \ref{eqs:loss}) and SC configurations have stayed the same as before.

\begin{table}[ht]
\begin{adjustwidth}{-.7in}{-.7in}
\begin{center}
\footnotesize
\setlength{\tabcolsep}{2.2pt}
\begin{tabular}{cccccccccc}
\toprule
\bf{$\tau$} & \bf{Ckpt.} & {\bf FLOPs} & {\bf GLUE} & {\bf SuperGLUE} & {\bf SQuAD} & {\bf CNNDM} & {\bf Rainbow} & {\bf MMLU} & {\bf BBH} \\ \midrule
0 & 500K & 1.0 & 88.92 $\pm$ 0.27 & 85.10 $\pm$ 0.43 & 91.30 $\pm$ 0.10 & 34.14 $\pm$ 0.02 & 81.48 $\pm$ 0.22 & 55.59 $\pm$ 0.84  & 40.30 $\pm$ 0.30 \\[0.4em]
120K & 500K & 1.06 & {\bf 89.66} $\pm$ 0.19 & {\bf 86.06} $\pm$ 0.47 & {\bf 91.36} $\pm$ 0.10 & {\bf 34.22} $\pm$ 0.18 & {\bf 82.68} $\pm$ 0.23 & {\bf 57.78} $\pm$ 1.01 & {\bf 42.07} $\pm$ 1.44 \\
\midrule
0 & 1M & 2.0 & 89.24 $\pm$ 0.17 & 86.11 $\pm$ 0.76 & 91.52 $\pm$ 0.04 & 34.24 $\pm$ 0.08 & 82.97 $\pm$ 0.20 & 58.72 $\pm$ 0.61 & 42.35 $\pm$ 0.72 \\[0.4em]
120K & 1M & 2.06 & {\bf 89.90} $\pm$ 0.26 & {\bf 86.38} $\pm$ 0.80 & {\bf 91.53} $\pm$ 0.13 & {\bf 34.27} $\pm$ 0.26 & {\bf 83.92} $\pm$ 0.32 & {\bf 59.06} $\pm$ 0.90 & {\bf 44.22} $\pm$ 1.52 \\
\bottomrule
\end{tabular}
\end{center}
\end{adjustwidth}
\vspace{2mm}
\caption{Average score of each downstream tasks for $\spactor_{\textrm{Large}}(\tau)$. $\tau=0$ corresponds to the baseline. The mean value and standard deviation across three independent runs. We bold numbers for $\spactor_{\textrm{Large}}(\tau)$ with a higher mean than baseline at the same pre-training steps. }
\label{tab:large_finetune_results}
\end{table}

Table \ref{tab:large_finetune_results} lists fine-tuning results for the same set of benchmarks as Base model. Because of the choice of generator $G$, the extra compute budget at 500K and 1M checkpoints is now 6\% and 3\% respectively. Just like previous experiments, we see that $\spactor_{\textrm{Large}}(\tau)$ consistently outperforms the baseline with a significant margin, measured by standard deviation. For GLUE, SuperGLUE and CNN/DailyMail, the 500K checkpoint of $\spactor_{\textrm{Large}}$ leads to better or equal downstream metrics compared to 1M checkpoint of baseline, while the rest of the tasks, the former is behind the latter, but the difference is within $1\sigma$. This results in an overall compute saving of 35\%. We conclude that \spactor\ method scales well as model size grows, probably because RTD provides purely complementary information on top of vanilla SC training objective. The breakdown of individual task is given in Appendix \ref{app:benchmark-breakdown}. 

\section{Related Work}
\citet{NIPS2015_7137debd, ramachandran-etal-2017-unsupervised} introduced language modeling with in-domain data to pre-train RNN sequence models. With the invention of transformer architecture \citep{NIPS2017_3f5ee243}, pre-training has become a standard paradigm to scale language models beyond O(100B) parameters, which often leads to strong performance on natural language tasks.

Assorted pre-training objectives have been studied in the literature, among which the most popular ones are causal language modeling (CLM) \citep{radford2018improving, radford2019language}, prefix language modeling (PLM) \citep{liu2018generating, T5}, masked language modeling (MLM) \citep{devlin-etal-2019-bert}. It has been understood that different pre-training objectives correlate with performance in different downstream tasks \citep{wang2022language}; therefore, one naturally curates  a mixtures of these objectives \citep{dong2019unified, tay2022unifying} such that the pre-trained LLM may inherit strength from them all.

Subsequent work also attempts to improve individual objectives. For MLM as an example, \citet{joshi2020spanbert} introduced \textsc{SpanBERT}, which masks contiguous tokens and uses span boundary to assist prediction. Inspired by that, \citet{T5, lewis-etal-2020-bart} considered a denoising objective where contiguous tokens are replaced with a single mask token, and showed that it achieves the best performances among other denoising options for encoder-decoder models.

The drawback of plain MLM, as well as other variants, is that not all tokens need to be attended to in order to figure out the ground truth. The existence of mask token \texttt{[M]} also creates misalignment between pre-train and downstream tasks. ELECTRA \citep{clark2020electra} rectifies those issues by jointly training a generator model that fills masked positions with plausible tokens, while the main model learning to detect which tokens have been replaced (\textit{i.e.} the RTD loss). The authors showed that ELECTRA significantly reduces the computing cost compared to other larger networks such as GPT \citep{radford2018improving} and XLNet \citep{NEURIPS2019_dc6a7e65}. Further extensions of ELECTRA can be found in \citet{meng2021cocolm, meng2022amos, he2021deberta, bajaj2022metro}.

Besides its success in BERT models, few works have attempted ELECTRA in T5. This is partially because RTD by itself is discriminative rather than generative in nature. As described in Section \ref{sec:method}, instead of \emph{replacing} SC with RTD, we \emph{combine} them to form a hybrid of pre-training objectives. The hybrid objective is evaluated on each individual input, where RTD learns a text representation while SC learns token generation. A closely related work that explored hybrid objective is PEGASUS \citep{zhang2020pegasus}; We emphasize our difference from PEGASUS in the following aspects: (\rn{1}) PEGASUS de-noises MLM in the encoder. For encoder component, RTD usually brings more benefit due to all token attention \citep{clark2020electra}; in addition, leaving MLM mask {\tt [M]} as model input hurts SC more, because token replacement can generate at least a proportion of context correctly; (\rn{2}) PEGASUS focuses exclusively on text summarization tasks.

Finally, there has been research on continued pre-training in LLMs, with focus on model adaptation: either adapting \emph{data} \citep{gururangan-etal-2020-dont}, or adapting the training \emph{objective} \citep{wang2022language} towards downstream tasks. The continued pre-training used in this paper is neither of the above two scenarios, rather it is more akin to curriculum type of training \citep{bengio2009curriculum, braun2017curriculum}: the difficulty of the objective changes as training progresses.

\section{Conclusion and Future Work}

In this paper, we construct a novel combination of pre-training objectives: span corruption (SC) \citep{T5} and replaced token detection (RTD) \citep{clark2020electra}, which enables the language model to learn from two signals simultaneously for every single input.  

In Section \ref{sec:introduction} and \ref{sec:experiments}, we argue empirically that RTD and SC cannot be co-trained for long durations since the downstream task performance would deteriorates sharply as pre-training progresses. It is then natural to propose a two-staged pre-training recipe, where after $\tau$ iterations we continue training with SC alone. We show that this approach is highly effective, where the model is able to reach the same performance as baseline with significantly less compute, while outperforming baseline given the same compute budget. Our observation also indicates that high quality data is critical for preserving and improving language abilities in later iterations.

There are a few limitations in the current scope of the paper. First, one may wonder whether a continuous pre-training curriculum exists. For example, smoothly varying the $\lambda_1$, $\lambda_2$ parameters, or MLM masking ratio. Secondly, our results are restricted to encoder-decoder architecture. It is interesting to extend the work to other architectures, and explore the scaling behavior along the lines of \citet{weiemergent, tay2022transcending}. We plan to leave those for future work.

\newpage
% References
\bibliographystyle{apalike}
\bibliography{main}

\newpage
\appendix
\onecolumn
\section{Training Hyperparameters}

In this section, we summarize more details of hyperparameter choices for both pre-training and fine-tuning. We only tune those parameters for $\spactor_{\textrm{Base}}(\tau)$ and then choose most of the optimal parameters for $\spactor_{\textrm{Large}}(\tau)$ experiments.

\subsection{Pre-training Hyperparameters}
\label{app:pre-training}

To select hyperparameters for T5-Base model, we run $\spactor_{\textrm{Base}}(\infty)$ with batch size 2048 to 250K steps, and then fine-tune the final checkpoints on a subset of downstream tasks (SuperGLUE, SQuAD) and select a set of reasonable values based on validation scores. For coefficients $\lambda_{1,2}$ in loss function, \textit{i.e.} Equation~\ref{eqs:loss}, we apply a simple grid search such that $\lambda_{1,2} \in \left[1.0, 10.0, 20.0, 50.0 \right]$. For the additional token level masking ratio, we experiment with $r_{\mathrm{MLM}} = \left[5\%, 10\%, 15\%, 20\%, 25\% \right]$ and find that a masking ratio of 15\% works the best. Indeed, a ratio that is too small would result in generator $G$ producing few different tokens from the initial input; while a ratio that is too large leads to an overly-corrupted input for the discriminator $D$, further affecting $D$ from training SC properly.

We also experiment with different generator architecture and sizes, in particular, selecting from encoder-only or encoder-decoder architecture. It is found that an encoder-only architecture suffices and there is no quality degradation using a linear projection layer mapping encoder output to the probability distribution of tokens in the vocabulary. We also compare final downstream performances when $G$ is a 3-layer, 4-layer, 6-layer model. Same as \citet{clark2020electra}, when $G$ is around 1/4 - 1/3 the size of the \emph{encoder} of $D$, the result is optimal.

The hyperparameter set is then fixed throughout the remainder of the empirical evaluation across all checkpoints and benchmarks. For T5-Large model, we re-use majority of the hyperparameters except scaling generator accordingly and increasing the MLM ratio from 15\% to 20\%.

\begin{table}[ht]
\begin{center}
\setlength{\tabcolsep}{15pt}
\begin{tabular}{lll}
\toprule
\bf{Parameter} & {\bf T5-Base Value} & {\bf T5-Large Value} \\ \midrule
Discriminator Layers & 12 & 24 \\[0.2em]
Discriminator Num Heads & 12 & 16 \\[0.2em]
Discriminator Hidden Dimension & 768 & 1024 \\[0.2em]
Discriminator MLP Size & 3072 & 4096 \\[0.2em]
RTD Head MLP Size & 3072 & 4096 \\[0.2em]
RTD Head MLP Activation & GELU & GELU \\[0.2em]
Generator Layers & 4 & 6 \\[0.2em]
Generator MLP Size & 1024 & 2048 \\[0.2em]
Input Length & 512 & 512 \\[0.2em]
Batch Size & 2048 & 2048 \\[0.2em]
Span Corruption & ($r$ = 15\%, $\mu$ = 3.0)  & ($r$ = 15\%, $\mu$ = 3.0) \\[0.2em]
MLM Ratio & 15\% & 20\% \\[0.2em]
Warmup Steps & $\kappa = 10,000$ & $\kappa = 10,000$ \\[0.2em]
Learning Rate Schedule & $1.0 / \sqrt{\max(n, \kappa)}$ & $1.0 / \sqrt{\max(n, \kappa)}$ \\[0.2em]
$(\lambda_1, \lambda_2)$ & $(10.0, 10.0)$ & $(10.0, 10.0)$ \\
\bottomrule
\end{tabular}
\end{center}
\vspace{2mm}
\caption{Model architecture and pre-training hyperparameters for \spactor. The RTD head uses the GELU activation initially proposed in \citep{hendrycks2016gaussian, shazeer2020glu}.}
\label{tab:pretrain_params}
\end{table}

\subsection{Fine-tuning Hyperparameters}
\label{app:fine-tuning}

For all the tasks except FLAN instruction-tuning, we fix a constant learning rate 1e-3, dropout rate 0.1, and batch size 128. For FLAN, we use constant learning rate 5e-4, dropout rate 0.05 and batch size 64, following \citet{chung2022scaling}. For the latter we also reset optimizer states since the data distribution is very different from pre-training corpus. We fine-tune for sufficiently long iterations, typically 250K - 300K, to ensure convergence.

Existing literature sometimes sweeps over a set of fine-tuning hyperparameters to get the optimal performance (for example, see \citet{aribandi2021ext5}). In our experiment, we found that the relative improvement stays the same regardless of finetuning hyperparameters that we searched, so we believe a fixed parameter set suffices to compare baseline against \spactor.

\section{Statistical Analysis of Validation Metric}
\label{app:stats-testing}

In Figure \ref{fig:pretraining-val-diff} we plot the validation loss difference between $\spactor_{\textrm{Base}}(\infty)$ evaluated with vanilla SC input text $X_{\mathrm{c}}$ or noisy input text $\widehat{X_{\mathrm{c}}}$ generated from $G$. We now perform linear regression and hypothesis testing and argue that the trend is significant.

We use simple linear regression where the $y$ variable is the cross-entropy difference, and the $x$ variable is training steps starting from 120K. We do not include data before 120K step because they have not reached the plateau. Let
\begin{equation}
    y = \beta_0 \cdot x + \beta_1
\end{equation}
to be the regression line. The hypothesis testing is
\begin{equation}
    H_0: \beta_0 = 0,  \ \ \ \ \ \  H_A: \beta_0 \neq 0.
\end{equation}

The regression results give $\beta_0 = -6.90 \times 10^{-5}$ with $p$-value $0.01$. We therefore conclude that the validation difference is trending with a high confidence to reject null hypothesis $H_0$.

\section{Average Score v.s. Pre-training FLOPs}
\label{app:score-flops-plot}

\begin{figure*}[t]
\centering  
\begin{subfigure}[b]{0.35\linewidth}
    \centering
    \includegraphics[width=\textwidth]{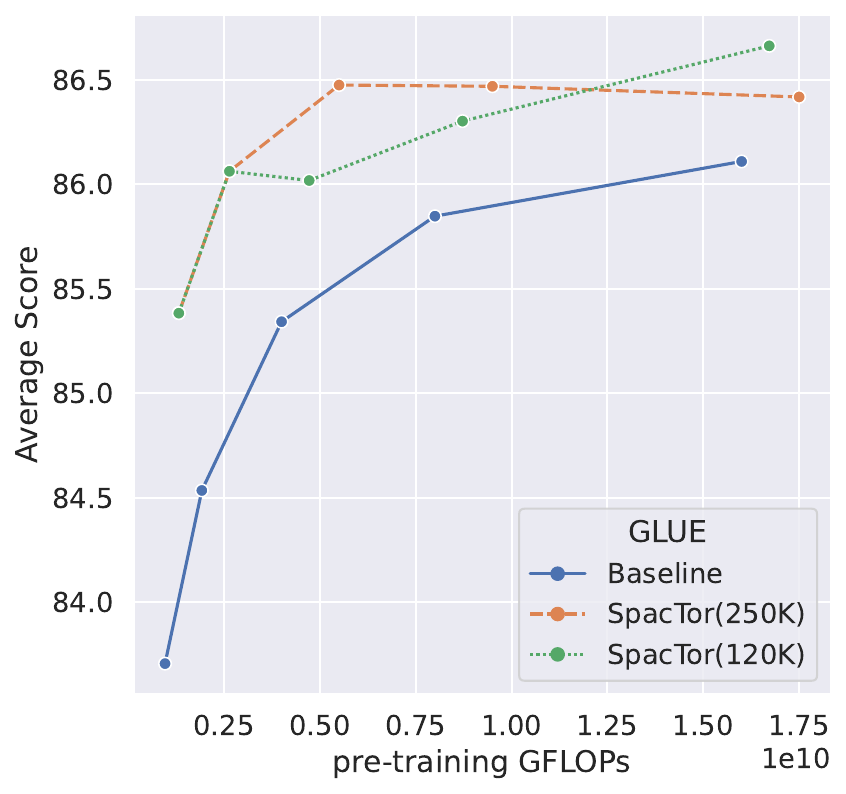}
    \caption{GLUE}
    \label{fig:glue-flops}
\end{subfigure}\hspace{10mm}
\begin{subfigure}[b]{0.35\linewidth}
    \centering
    \includegraphics[width=\textwidth]{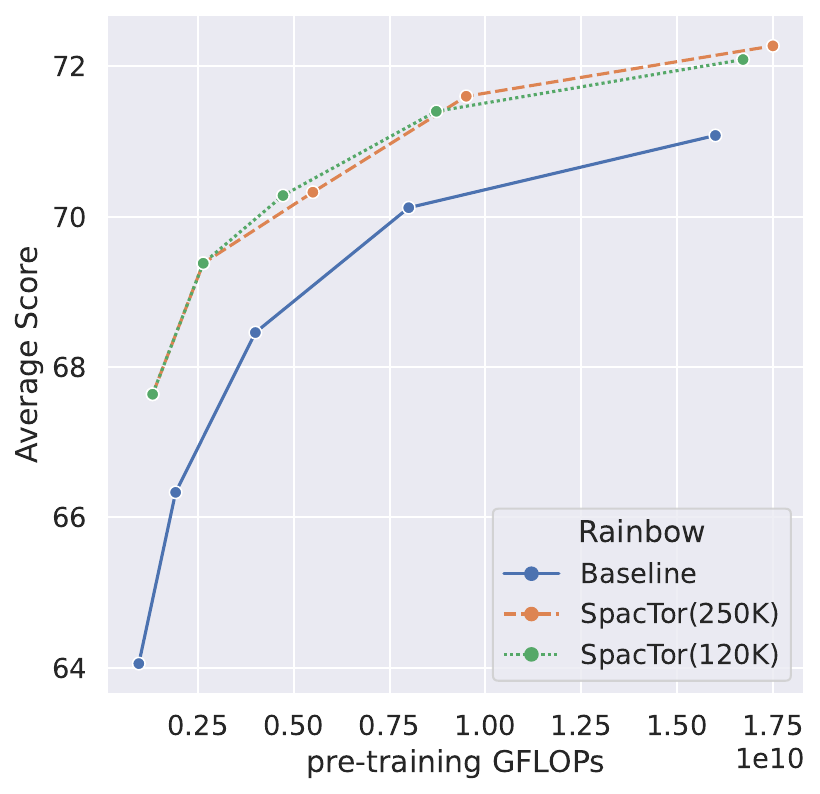}
    \caption{Rainbow}
    \label{fig:rainbow-flops}
\end{subfigure}
\begin{subfigure}[b]{0.35\linewidth}
    \centering
    \includegraphics[width=\textwidth]{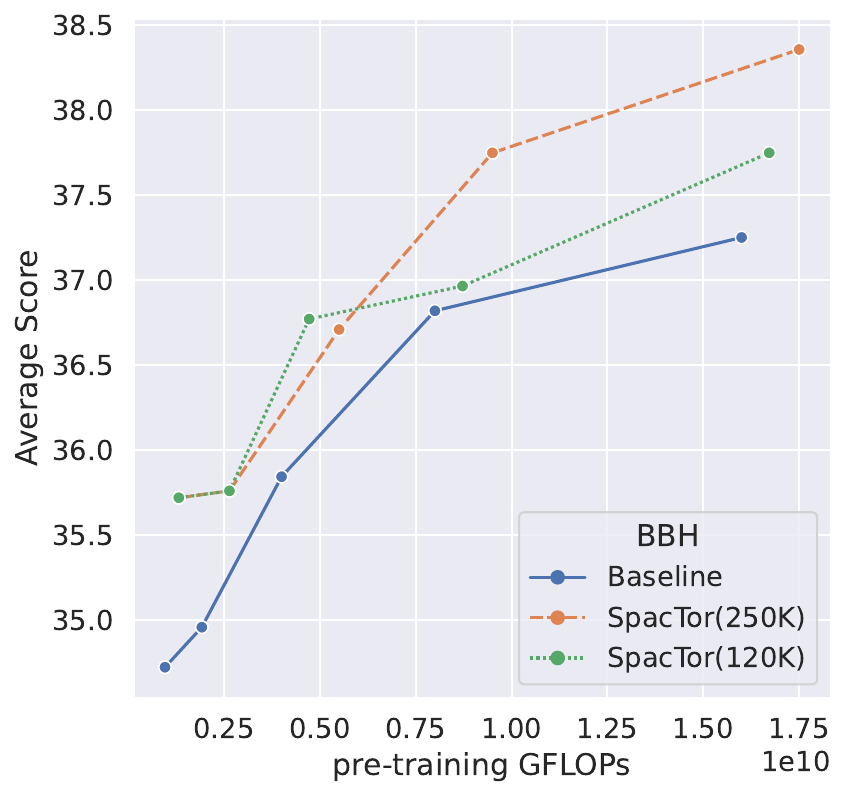}
    \caption{BBH}
    \label{fig:bbh-flops}
\end{subfigure}\hspace{10mm}
\begin{subfigure}[b]{0.35\linewidth}
    \centering
    \includegraphics[width=\textwidth]{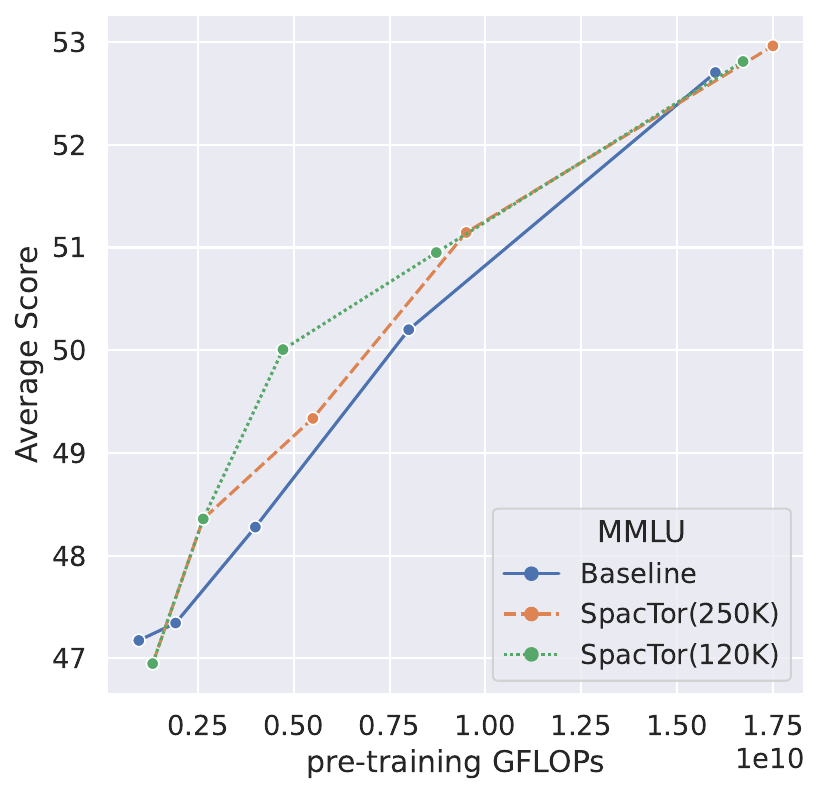}
    \caption{MMLU}
    \label{fig:mmlu-flops}
\end{subfigure}
\caption{\spactor\ performances on GLUE, Rainbow, BBH and MMLU with respect to pre-training FLOPs for T5-Base model.}
\label{fig:spactor-flops-remaining}
\end{figure*}

In this section, we include the rest of average score versus pre-training FLOPs plots in Figure~\ref{fig:spactor-flops-remaining} for $\spactor_{\textrm{Base}}$ models. While for MMLU the gain is smaller (around 20\% FLOPs saving) and $\spactor_{\textrm{Base}}$ gradually converge to baseline, the other tasks we still see substantial gains. On the other hand, whether $\spactor_{\textrm{Base}}(120\textrm{K})$ is better than $\spactor_{\textrm{Base}}(250\textrm{K})$ is undetermined because the comparison is task dependent. This implies that the some tasks benefit more from longer training on the hybrid objective (Equation~\ref{eqs:loss}).

\section{Detailed Eval Metrics}
\label{app:benchmark-breakdown}

In this section, we give detailed metric breakdown of individual sub-tasks for $\spactor_{\textrm{Base}}$ and $\spactor_{\textrm{Large}}$, as a supplement to aggregated results of Table \ref{tab:finetune_results} and Table \ref{tab:large_finetune_results}.

\begin{table}[ht]
\begin{adjustwidth}{-.7in}{-.7in}
\begin{center}
\small
\begin{tabular}{ccccccccccc}
\toprule
\bf{Size} & \bf{$\tau$} & {\bf Ckpt} & {\bf CoLA} & {\bf MNLI} & {\bf MRPC} & {\bf QNLI} & {\bf QQP} & {\bf RTE} & {\bf SST-2} & {\bf STS-B} \\ \midrule
\multirow{6}{*}{Base} & 0 & 500K & 55.50 & 88.22 / 88.11 & 93.02 / 90.44 & 92.99 & 89.41 / 92.12 & 81.95 & 94.61 & 90.05 / 89.91 \\[0.2em]
& 250K & 500K & {\bf 58.20} & {\bf 88.35 / 88.20} & {\bf 93.36 / 90.69} & {\bf 93.10} & 89.36 / 92.08 & {\bf 83.75} & 94.50 & {\bf 90.34 / 90.09} \\[0.2em]
& 120K & 500K & {\bf 57.02} & {\bf 88.34 / 88.19} & 92.77 / 89.95 & {\bf 93.04} & 89.33 / 92.06 & {\bf 82.67} & {\bf 94.72} & {\bf 90.59 / 90.40} \\[0.2em]
& 0 & 1M & 54.73 & 88.40 / 88.40 & 92.96 / 90.20 & 93.12 & 89.44 / 92.15 & 82.67 & 94.84 & 90.69 / 90.45 \\[0.2em]
& 250K & 1M & {\bf 55.98} & {\bf 88.46 / 88.45} & {\bf 93.17 / 90.69} & {\bf 93.23} & 89.41 / 92.09 & {\bf 83.75} & 94.72 & 90.62 / {\bf 90.48} \\[0.2em]
& 120K & 1M & {\bf 58.06} & {\bf 88.50 / 88.53} & {\bf 93.33 / 90.69} & {\bf 93.26} & 89.41 / 92.12 & {\bf 84.12} & 94.72 & 90.48 / 90.41 \\
\bottomrule
\end{tabular}
\end{center}
\end{adjustwidth}
\vspace{2mm}
\caption{Breakdown of GLUE \citep{wang2018glue} sub-tasks for $\spactor_{\textrm{Base}}$. Each score corresponds to the median of 5 independent runs. The metrics for each sub-task are: Matthews correlation coefficient for CoLA, matched/mismatched accuracy for MNLI, F1/accuracy for MRPC, accuracy for QNLI, F1/accuracy for QQP, accuracy for RTE, accuracy for SST-2, Pearson correlation coefficient/Spearman correlation coefficient for STS-B.}
\label{tab:glue_breakdown}
\end{table}

\begin{table}[ht]
\begin{adjustwidth}{-.7in}{-.7in}
\begin{center}
\small
\begin{tabular}{ccccccccccc}
\toprule
\bf{Size} & \bf{$\tau$} & {\bf Ckpt} & {\bf CoLA} & {\bf MNLI} & {\bf MRPC} & {\bf QNLI} & {\bf QQP} & {\bf RTE} & {\bf SST-2} & {\bf STS-B} \\ \midrule
\multirow{4}{*}{Large} & 0 & 500K & 64.14 & 90.29 / 90.39 & 93.66 / 91.18 & 94.82 & 90.13 / 92.62 & 89.17 & 95.76 & 91.85 / 91.64 \\[0.2em]
& 120K & 500K & {\bf 68.28} & {\bf 90.55 / 90.57} & {\bf 94.33 / 92.16} & {\bf 94.93} & 90.06 / {\bf 92.63} & {\bf 90.25} & {\bf 96.22} & 91.78 / 91.54 \\[0.2em]
& 0 & 1M & 63.33 & 90.80 / 90.94 & 94.16 / 91.91 & 95.04 & 90.20 / 92.68 & 90.61 & 96.10 & 91.83 / 91.72 \\[0.2em]
& 120K & 1M & {\bf 67.43} & {\bf 90.93 / 90.95} & {\bf 94.41 / 92.16} & {\bf 95.20} & {\bf 90.25 / 92.70} & {\bf 91.34} & {\bf 96.44} & {\bf 92.17 / 92.00}  \\
\bottomrule
\end{tabular}
\end{center}
\end{adjustwidth}
\vspace{2mm}
\caption{Breakdown of GLUE \citep{wang2018glue} sub-tasks for $\spactor_{\textrm{Large}}$. Each score corresponds to the median of 3 independent runs. The metrics for each sub-task are: Matthews correlation coefficient for CoLA, matched/mismatched accuracy for MNLI, F1/accuracy for MRPC, accuracy for QNLI, F1/accuracy for QQP, accuracy for RTE, accuracy for SST-2, Pearson correlation coefficient/Spearman correlation coefficient for STS-B.}
\label{tab:glue_breakdown_large}
\end{table}

\begin{table}[ht]
\begin{adjustwidth}{-.7in}{-.7in}
\begin{center}
\small
\begin{tabular}{ccccccccccc}
\toprule
\bf{Size} & \bf{$\tau$} & {\bf Ckpt} & {\bf BoolQ} & {\bf CB} & {\bf COPA} & {\bf MultiRC} & {\bf ReCoRD} & {\bf RTE} & {\bf WiC} & {\bf WSC} \\ \midrule
\multirow{6}{*}{Base} & 0 & 500K & 81.99 & 96.07 / 96.43 & 70.00 & 76.15 / 37.88 & 77.65 / 78.53 & 81.59 & 68.97 & 83.65 \\[0.2em]
& 250K & 500K & {\bf 82.32} & 93.70 / 94.64 & {\bf 73.00} & {\bf 77.09 / 40.19} & 77.65 / {\bf 78.55} & {\bf 83.03} & {\bf 69.44} & {\bf 85.58} \\[0.2em]
& 120K & 500K & {\bf 82.72} & 95.03 / 96.43 & {\bf 74.00} & {\bf 77.04 / 38.93} & {\bf 77.92 / 78.92} & {\bf 82.31} & {\bf 70.22} & {\bf 84.62}\\[0.2em]
& 0 & 1M & 82.39 & 97.36 / 96.43 & 72.00 & 77.10 / 39.66 & 78.10 / 79.10 & 83.03 & 69.44 & 86.54 \\[0.2em]
& 250K & 1M & {\bf 82.78} & 91.89 / 94.64 & {\bf 76.00} & {\bf 77.63 / 41.03} & 78.05 / 79.03 & 83.03 & 69.12 & 85.58 \\[0.2em]
& 120K & 1M & {\bf 82.66} & 95.04 / 94.64 & {\bf 74.00} & {\bf 77.94 / 41.76} & {\bf 78.21 / 79.20} & 82.67 & 69.28 & 82.69\\
\bottomrule
\end{tabular}
\end{center}
\end{adjustwidth}
\vspace{2mm}
\caption{Breakdown of SuperGLUE \citep{wang2019superglue} sub-tasks for $\spactor_{\textrm{Base}}$. Each score corresponds to the median of 5 independent runs. The metrics for each sub-task are: accuracy for BoolQ, average F1/accuracy for CB, accuracy for COPA, F1/Exact Match (EM) for MultiRC, EM/F1 for ReCoRD, accuracy for RTE, accuracy for RTE, WiC and WSC.}
\label{tab:superglue_breakdown}
\end{table}

\begin{table}[ht]
\begin{adjustwidth}{-.7in}{-.7in}
\begin{center}
\small
\begin{tabular}{ccccccccccc}
\toprule
\bf{Size} & \bf{$\tau$} & {\bf Ckpt} & {\bf BoolQ} & {\bf CB} & {\bf COPA} & {\bf MultiRC} & {\bf ReCoRD} & {\bf RTE} & {\bf WiC} & {\bf WSC} \\ \midrule
\multirow{4}{*}{Large} & 0 & 500K & 87.49 & 95.59 / 98.21 & 85.00 & 83.97 / 52.93 & 86.30 / 87.21 & 89.89 & 72.88 & 94.23\\[0.2em]
& 120K & 500K & 87.43 & {\bf 100.00 / 100.00} & {\bf 87.00} & {\bf 84.33 / 54.98} & {\bf 86.86 / 87.74} & {\bf 91.94} & {\bf 74.92} & 93.27\\[0.2em]
& 0 & 1M & 87.92 & 96.23 / 98.21 & 90.00 & 85.19 / 56.45 & 87.31 / 88.17 & 90.61 & 74.92 & 91.35 \\[0.2em]
& 120K & 1M & 87.80 & {\bf 100.00 / 100.00} & 87.00 & 85.05 / 56.35 & {\bf 87.43 / 88.29} & {\bf 90.97} & 73.98 & {\bf 93.27} \\
\bottomrule
\end{tabular}
\end{center}
\end{adjustwidth}
\vspace{2mm}
\caption{Breakdown of SuperGLUE \citep{wang2019superglue} sub-tasks for $\spactor_{\textrm{Large}}$. Each score corresponds to the median of 3 independent runs. The metrics for each sub-task are: accuracy for BoolQ, average F1/accuracy for CB, accuracy for COPA, F1/Exact Match (EM) for MultiRC, EM/F1 for ReCoRD, accuracy for RTE, accuracy for RTE, WiC and WSC.}
\label{tab:superglue_breakdown_large}
\end{table}

\begin{table}[ht]
\begin{adjustwidth}{-.7in}{-.7in}
\begin{center}
\footnotesize
\setlength{\tabcolsep}{3.0pt}
\begin{tabular}{ccccccccccc}
\toprule
\bf{Size} & \bf{$\tau$} & {\bf Ckpt} & {\bf SQuAD} & {\bf CNNDM} & {\bf $\alpha$NLI} & {\bf CosmosQA} & {\bf HellaSWAG} & {\bf PIQA} & {\bf SocialIQA} & {\bf WinoGrande} \\ \midrule
\multirow{6}{*}{Base} & 0 & 500K & 85.01 / 92.20 & 41.48 / 19.43 / 38.98 & 71.28 & 74.51 & 62.47 & 76.28 & 69.60 & 66.85 \\[0.2em]
& 250K & 500K & {\bf 85.34 / 92.42} & {\bf 41.50} / 19.41 / {\bf 39.03} & {\bf 71.87} & {\bf 75.58} & {\bf 66.80} & {\bf 76.93} & {\bf 70.32} & {\bf 67.96} \\[0.2em]
& 120K & 500K & {\bf 85.37 / 92.42} & {\bf 41.54 / 19.47 / 39.04} & {\bf 71.67} & {\bf 75.08} & {\bf 66.11} & {\bf 77.15} & {\bf 70.01} & {\bf 68.03} \\[0.2em]
& 0 & 1M & 85.37 / 92.34 & 41.51 / 19.48 / 39.02 & 71.34 & 75.31 & 64.44 & 76.55 & 70.83 & 67.17 \\[0.2em]
& 250K & 1M & {\bf 85.63 / 92.57} & {\bf 41.69 / 19.55 / 39.17} & {\bf 72.26} & {\bf 76.55} & {\bf 67.19} & {\bf 77.42} & 70.78 & {\bf 68.35} \\[0.2em]
& 120K & 1M & {\bf 85.54 / 92.58} & {\bf 41.76 / 19.60 / 39.24} & {\bf 72.39} & {\bf 76.58} & {\bf 67.30} & {\bf 77.20} & {\bf 71.29} & {\bf 68.67} \\
\bottomrule
\end{tabular}
\end{center}
\end{adjustwidth}
\vspace{2mm}
\caption{Breakdown of SQuAD \citep{rajpurkar2016squad}, CNN/DailyMail \citep{hermann2015teaching} and Rainbow \citep{lourie2021unicorn} sub-tasks for $\spactor_{\textrm{Base}}$. Each score corresponds to the median of 5 independent runs. The metrics for each sub-task are: EM/F1 for SQuAD, Rouge-1/Rouge-2/Rouge-L for CNN/DailyMail, and accuracy for all the Rainbow tasks.}
\label{tab:squad-cnndm-rainbow_breakdown}
\end{table}

\begin{table}[ht]
\begin{adjustwidth}{-.7in}{-.7in}
\begin{center}
\footnotesize
\setlength{\tabcolsep}{3.0pt}
\begin{tabular}{ccccccccccc}
\toprule
\bf{Size} & \bf{$\tau$} & {\bf Ckpt} & {\bf SQuAD} & {\bf CNNDM} & {\bf $\alpha$NLI} & {\bf CosmosQA} & {\bf HellaSWAG} & {\bf PIQA} & {\bf SocialIQA} & {\bf WinoGrande} \\ \midrule
\multirow{4}{*}{Large} & 0 & 500K & 88.13 / 94.52 & 42.39 / 20.23 / 39.82 & 80.29 & 83.62 & 85.28 & 83.41 & 77.23 & 79.56\\[0.2em]
& 120K & 500K & {\bf 88.18 / 94.54} & {\bf 42.47 / 20.25 / 39.95} & {\bf 82.11} & {\bf 84.32} & {\bf 87.25} & {\bf 84.44} & {\bf 77.74} & {\bf 80.19}\\[0.2em]
& 0 & 1M & 88.37 / 94.67 & 42.56 / 20.23 / 40.00 & 82.05 & 84.66 & 87.01 & 84.28 & 78.25 & 81.29 \\[0.2em]
& 120K & 1M & {\bf 88.37} / 94.66 & 42.47 / {\bf 20.32} / 39.95 & {\bf 83.29} & {\bf 86.06} & {\bf 88.56} & {\bf 85.75} & {\bf 78.61} & {\bf 81.93}\\
\bottomrule
\end{tabular}
\end{center}
\end{adjustwidth}
\vspace{2mm}
\caption{Breakdown of SQuAD \citep{rajpurkar2016squad}, CNN/DailyMail \citep{hermann2015teaching} and Rainbow \citep{lourie2021unicorn} sub-tasks for $\spactor_{\textrm{Large}}$. Each score corresponds to the median of 5 independent runs. The metrics for each sub-task are: EM/F1 for SQuAD, Rouge-1/Rouge-2/Rouge-L for CNN/DailyMail, and accuracy for all the Rainbow tasks.}
\label{tab:squad-cnndm-rainbow_breakdown_large}
\end{table}

\begin{table}[ht]
\begin{adjustwidth}{-.7in}{-.7in}
\begin{center}
\small
\setlength{\tabcolsep}{2.7pt}
\begin{tabular}{ccccccccccc}
\toprule
\bf{Size} & \bf{$\tau$} & {\bf Ckpt} &  \thead{{\bf Boolean}\\ {\bf Expressions}} &  \thead{{\bf Causal}\\ {\bf Judgement}} &  \thead{{\bf Date}\\ {\bf Understanding}} &  \thead{{\bf Disambi-}\\ {\bf guation QA}} &  \thead{{\bf Dyck}\\ {\bf Languages}} &  \thead{{\bf Formal}\\ {\bf Fallacies}} & \thead{{\bf Geometric}\\ {\bf Shapes}} & \thead{{\bf Hyper-}\\ {\bf baton}} \\
\midrule
\multirow{6}{*}{Base} & 0 & 500K & 57.60 & 56.68 & 38.00 & 52.00 & 6.80 & 56.00 & 22.40 & 68.80\\[0.2em]
& 250K & 500K & 55.20 & 55.61 & {\bf 40.40} & {\bf 60.00} & 3.60 & {\bf 58.80} & {\bf 28.40} & 66.40\\[0.2em]
& 120K & 500K & 54.40 & 55.61 & {\bf 42.80} & {\bf 56.80} & 6.40 & {\bf 59.60} & 21.20 & 65.20 \\[0.2em]
& 0 & 1M & 54.80 & 55.61 & 40.80 & 58.40 & 6.40 & 60.40 & 18.00 & 72.00 \\[0.2em]
& 250K & 1M & {\bf 59.60} & 54.55 & {\bf 44.00} & {\bf 60.80} & 5.20 & 60.00 & {\bf 29.60} & 69.20\\[0.2em]
& 120K & 1M & {\bf 58.00} & {\bf 56.15} & 40.00 & {\bf 60.80} & 3.60 & 60.00 & {\bf 30.00} & 62.00\\
\midrule \midrule
\bf{Size} & \bf{$\tau$} & {\bf Ckpt} & \thead{{\bf Logical}\\ {\bf Deduction} \\ {\bf 5 Objects}} &  \thead{{\bf Logical}\\ {\bf Deduction} \\ {\bf 7 Objects}} &  \thead{{\bf Logical}\\ {\bf Deduction} \\ {\bf 3 Objects}} & \thead{{\bf  Movie Reco-}\\ {\bf mmendation}} &  \thead{{\bf Multistep}\\ {\bf Arithmetic}\\ {\bf Two}} &  {\bf Navigate} &  \thead{{\bf Object}\\ {\bf Counting}} &  \thead{{\bf Penguins}\\ {\bf In A}\\ {\bf Table}} \\
\midrule
\multirow{6}{*}{Base} & 0 & 500K & 28.00 & 26.40 & 42.40 & 49.20 & 1.60 & 64.00 & 24.80 & 32.88 \\[0.2em]
& 250K & 500K & {\bf 32.40} & {\bf 29.20} & {\bf 44.00} & 47.60 & 1.20 & {\bf 64.40} & 23.20 & 27.40 \\[0.2em]
& 120K & 500K & {\bf 31.20} & {\bf 29.60} & {\bf 45.60} & 47.20 & 1.20 & 64.00 & {\bf 26.40} & 28.77 \\[0.2em]
& 0 & 1M & 33.20 & 25.60 & 44.40 & 47.20 & 1.20 & 64.00 & 28.00 & 32.88 \\[0.2em]
& 250K & 1M & 32.40 & {\bf 32.40} & {\bf 47.60} & {\bf 47.60} & {\bf 1.60} & {\bf 64.80} & {\bf 28.40} & 30.82 \\[0.2em]
& 120K & 1M & {\bf 34.00} & {\bf 29.60} & {\bf 47.20} & {\bf 47.60} & 1.20 & {\bf 65.60} & {\bf 29.60} & 30.14\\
\midrule \midrule
\bf{Size} & \bf{$\tau$} & {\bf Ckpt} & \thead{{\bf Reasoning}\\ {\bf About}\\ {\bf Colored}\\ {\bf Objects}} &  \thead{{\bf Ruin}\\ {\bf Names}} &  \thead{{\bf Salient}\\ {\bf Translation}\\ {\bf Error}\\ {\bf Detection}} & {\bf Snarks} &  \thead{{\bf Sports}\\ {\bf Under-}\\ {\bf standing}} &  \thead{{\bf Temporal}\\ {\bf Sequences}} & \thead{{\bf Tracking}\\ {\bf Shuffled}\\ {\bf Objects} \\ {\bf 5 Objects}} &  \thead{{\bf Tracking}\\ {\bf Shuffled}\\ {\bf Objects}\\ {\bf 7 Objects}}\\
\midrule
\multirow{6}{*}{Base} & 0 & 500K & 4.80 & 28.00 & 26.00 &  53.93 & 56.80 & 28.40 & 22.40 & 18.00 \\[0.2em]
& 250K & 500K & {\bf 34.80} & 28.00 & {\bf 27.20} & 53.93 & 56.80 & {\bf 28.80} & 22.00 & 17.20\\[0.2em]
& 120K & 500K & {\bf 32.80} & 28.00 & 26.00 & 53.93 & 56.40 & {\bf 28.80} & 22.40 & 18.00 \\[0.2em]
& 0 & 1M & 35.20 & 28.00 & 30.00 &  54.49 & 57.20 & 30.40 & 21.20 & 18.40 \\[0.2em]
& 250K & 1M & 32.00 & 28.00 & 29.20 & 53.93 & {\bf 57.60} & 26.40 & 21.20 & 18.40\\[0.2em]
& 120K & 1M & 34.80 & 28.00 & {\bf 30.40} & 53.93 & {\bf 57.60} & 30.00 & {\bf 21.60} & 17.20\\
\midrule \midrule
\bf{Size} & \bf{$\tau$} & {\bf Ckpt} & \thead{{\bf Tracking}\\ {\bf Shuffled}\\ {\bf Objects}\\ {\bf 3 Objects}} &  \thead{{\bf Web Of}\\ {\bf Lies}} &  \thead{{\bf Word}\\ {\bf Sorting}} & & & & & \\
\midrule
\multirow{6}{*}{Base} & 0 & 500K & 37.60 & 55.60 & 4.00 & & & & & \\[0.2em]
& 250K & 500K & 36.40 & 54.40 & 4.00 & & & & & \\[0.2em]
& 120K & 500K & 35.20 & {\bf 56.80} & 4.00 & & & & & \\[0.2em]
& 0 & 1M & 38.00 & 55.20 & 4.00\\[0.2em]
& 250K & 1M & 36.00 & 54.40 & 4.00\\[0.2em]
& 120K & 1M & 35.60 & 54.80 & 4.00\\
\bottomrule
\end{tabular}
\end{center}
\end{adjustwidth}
\vspace{2mm}
\caption{Breakdown of 27 BBH \citep{srivastava2022beyond} tasks with direct answers for $\spactor_{\textrm{Base}}$. The metric are all accuracy.}
\label{tab:bbh_breakdown}
\end{table}

\begin{table}[ht]
\begin{adjustwidth}{-.7in}{-.7in}
\begin{center}
\small
\setlength{\tabcolsep}{2.7pt}
\begin{tabular}{ccccccccccc}
\toprule
\bf{Size} & \bf{$\tau$} & {\bf Ckpt} &  \thead{{\bf Boolean}\\ {\bf Expressions}} &  \thead{{\bf Causal}\\ {\bf Judgement}} &  \thead{{\bf Date}\\ {\bf Understanding}} &  \thead{{\bf Disambi-}\\ {\bf guation QA}} &  \thead{{\bf Dyck}\\ {\bf Languages}} &  \thead{{\bf Formal}\\ {\bf Fallacies}} & \thead{{\bf Geometric}\\ {\bf Shapes}} & \thead{{\bf Hyper-}\\ {\bf baton}} \\
\midrule
\multirow{4}{*}{Large} & 0 & 500K & 59.60 & 59.36 & 47.60 & 67.20 & 2.40 & 59.20 & 14.80 & 72.00\\[0.2em]
& 120K & 500K & {\bf 60.80} & {\bf 59.36} & {\bf 50.80} & 66.40 & {\bf 6.00} & 56.80 & {\bf 32.40} & 67.20 \\[0.2em]
& 0 & 1M & 63.60 & 61.50 & 54.00 & 66.80 & 2.40 & 60.40 & 27.60 & 73.60 \\[0.2em]
& 120K & 1M & {\bf 64.40} & 59.89 & {\bf 54.40} & {\bf 67.60} & {\bf 4.80} & 57.20 & 24.80 & {\bf 85.60} \\
\midrule \midrule
\bf{Size} & \bf{$\tau$} & {\bf Ckpt} & \thead{{\bf Logical}\\ {\bf Deduction} \\ {\bf 5 Objects}} &  \thead{{\bf Logical}\\ {\bf Deduction} \\ {\bf 7 Objects}} &  \thead{{\bf Logical}\\ {\bf Deduction} \\ {\bf 3 Objects}} & \thead{{\bf  Movie Reco-}\\ {\bf mmendation}} &  \thead{{\bf Multistep}\\ {\bf Arithmetic}\\ {\bf Two}} &  {\bf Navigate} &  \thead{{\bf Object}\\ {\bf Counting}} &  \thead{{\bf Penguins}\\ {\bf In A}\\ {\bf Table}} \\
\midrule
\multirow{4}{*}{Large} & 0 & 500K & 44.00 & 46.00 & 56.00 & 60.80 & 1.20 & 59.60 & 40.00 & 34.93  \\[0.2em]
& 120K & 500K & {\bf 48.00} & {\bf 50.40} & {\bf 58.40} & {\bf 62.00} & {\bf 1.60} & {\bf 60.40} & 36.80 & {\bf 35.62} \\[0.2em]
& 0 & 1M & 44.40 & 48.80 & 61.60 & 54.80 & 1.20 & 62.40 & 42.40 & 39.04 \\[0.2em]
& 120K & 1M & {\bf 52.00} & {\bf 55.60} & {\bf 68.80} & {\bf 62.00} & {\bf 1.20} & {\bf 65.20} & 37.60 & {\bf 43.15} \\
\midrule \midrule
\bf{Size} & \bf{$\tau$} & {\bf Ckpt} & \thead{{\bf Reasoning}\\ {\bf About}\\ {\bf Colored}\\ {\bf Objects}} &  \thead{{\bf Ruin}\\ {\bf Names}} &  \thead{{\bf Salient}\\ {\bf Translation}\\ {\bf Error}\\ {\bf Detection}} & {\bf Snarks} &  \thead{{\bf Sports}\\ {\bf Under-}\\ {\bf standing}} &  \thead{{\bf Temporal}\\ {\bf Sequences}} & \thead{{\bf Tracking}\\ {\bf Shuffled}\\ {\bf Objects} \\ {\bf 5 Objects}} &  \thead{{\bf Tracking}\\ {\bf Shuffled}\\ {\bf Objects}\\ {\bf 7 Objects}}\\
\midrule
\multirow{4}{*}{Large} & 0 & 500K & 41.60 & 20.00 & 34.40 & 53.37 & 58.40 & 26.80 & 17.60 & 15.60\\[0.2em]
& 120K & 500K & {\bf 44.40} & {\bf 25.60} & 28.00 & {\bf 55.06} & {\bf 59.20} & {\bf 37.20} & {\bf 19.20} & 14.80 \\[0.2em]
& 0 & 1M & 44.80 & 25.60 & 34.80 & 58.99 & 58.80 & 30.40 & 17.60 & 14.40 \\[0.2em]
& 120K & 1M & {\bf 46.40} & 24.80 & {\bf 41.20} & 52.25 & 57.60 & {\bf 36.00} & {\bf 19.20} & {\bf 14.80} \\
\midrule \midrule
\bf{Size} & \bf{$\tau$} & {\bf Ckpt} & \thead{{\bf Tracking}\\ {\bf Shuffled}\\ {\bf Objects}\\ {\bf 3 Objects}} &  \thead{{\bf Web Of}\\ {\bf Lies}} &  \thead{{\bf Word}\\ {\bf Sorting}} & & & & & \\
\midrule
\multirow{4}{*}{Large} & 0 & 500K & 34.80 & 54.00 & 4.80 \\[0.2em]
& 120K & 500K & {\bf 35.20} & 53.60 & 4.00 \\[0.2em]
& 0 & 1M & 34.00 & 53.60 & 6.00 \\[0.2em]
& 120K & 1M & 33.60 & {\bf 58.00} & {\bf 6.00} \\
\bottomrule
\end{tabular}
\end{center}
\end{adjustwidth}
\vspace{2mm}
\caption{Breakdown of 27 BBH \citep{srivastava2022beyond} tasks with direct answers for $\spactor_{\textrm{Large}}$. The metric are all accuracy.}
\label{tab:bbh_breakdown_large}
\end{table}

\begin{table}[ht]
\begin{adjustwidth}{-.7in}{-.7in}
\begin{center}
\small
\setlength{\tabcolsep}{3.0pt}
\begin{tabular}{ccccccccccc}
\toprule
\bf{Size} & \bf{$\tau$} & {\bf Ckpt} &  \thead{{\bf Abstract}\\ {\bf Algebra}} &  \thead{{\bf Anatomy}} &  \thead{\bf Astronomy} &  \thead{{\bf Business}\\ {\bf Ethics}} &  \thead{{\bf Clinical}\\ {\bf Knowledge}} &  \thead{{\bf College}\\ {\bf Biology}} &  \thead{{\bf College}\\ {\bf Chemistry}} &  \thead{{\bf College}\\ {\bf Computer}\\  {\bf Science}}  \\
\midrule
\multirow{6}{*}{Base} & 0 & 500K & 36.36 & 50.00 & 50.00 & 63.64 & 55.17 & 50.00 & 50.00 & 63.64 \\[0.2em]
& 250K & 500K & 36.36 & 50.00 & {\bf 56.25} & 63.64 & 44.83 & 50.00 & {\bf 62.50} & 63.64\\[0.2em]
& 120K & 500K & 36.36 & 50.00 & 50.00 & {\bf 72.73} & 48.28 & 43.75 & {\bf 62.50} & 54.55\\[0.2em]
& 0 & 1M & 45.45 & 57.14 & 50.00 & 72.73 & 55.17 & 43.75 & 62.50 & 72.73 \\[0.2em]
& 250K & 1M & 45.45 & 50.00 & 50.00 & 63.64 & 51.72 & {\bf 50.00} & 62.50 & 63.64\\[0.2em]
& 120K & 1M & 36.36 & 57.14 & {\bf 56.25} & 72.73 & 51.72 & 43.75 & 62.50 & 63.64\\
\midrule \midrule
\bf{Size} & \bf{$\tau$} & {\bf Ckpt} &  \thead{{\bf College}\\ {\bf Mathematics}} & \thead{{\bf College}\\ {\bf Medicine}} &  \thead{{\bf College}\\ {\bf Physics}} &  \thead{{\bf Computer}\\ {\bf Security}} &  \thead{{\bf Conceptual}\\ {\bf Physics}} &  \thead{{\bf Econo-}\\ {\bf metrics}} &  \thead{{\bf Electrical}\\ {\bf Engineering}} & \thead{{\bf  Elementary}\\ {\bf Mathematics}} \\
\midrule
\multirow{6}{*}{Base} & 0 & 500K & 36.36 & 63.64 & 72.73 & 45.45 & 42.31 & 58.33 & 50.00 & 36.59 \\[0.2em]
& 250K & 500K & {\bf 45.45} & 63.64 & 72.73 & {\bf 63.64} & {\bf 46.15} & 50.00 & 43.75 & 36.59\\[0.2em]
& 120K & 500K & {\bf 45.45} & 63.64 & 63.64 & {\bf 54.55} & 42.31 & 50.00 & 50.00 & {\bf 39.02} \\[0.2em]
& 0 & 1M & 45.45 & 63.64 & 81.82 & 54.55 &  38.46 & 58.33 & 56.25 & 36.59\\[0.2em]
& 250K & 1M & 45.45 & 59.09 & 72.73 & {\bf 63.64} & 38.46 & 50.00 & 50.00 & {\bf 39.02}\\[0.2em]
& 120K & 1M & 45.45 & 63.64 & 63.64 & 54.55 & {\bf 42.31} & 50.00 & 50.00 & 34.15\\
\midrule \midrule
\bf{Size} & \bf{$\tau$} & {\bf Ckpt} & \thead{{\bf Formal}\\ {\bf Logic}} & \thead{{\bf Global}\\ {\bf Facts}} & \thead{{\bf High}\\ {\bf School}\\  {\bf Biology}} & \thead{{\bf High}\\ {\bf School}\\  {\bf Chemistry}} & \thead{{\bf High}\\ {\bf School}\\  {\bf Computer}\\ {\bf Science}} & \thead{{\bf High}\\ {\bf School}\\  {\bf European}\\ {\bf History}} & \thead{{\bf High}\\ {\bf School}\\  {\bf Geography}} & \thead{{\bf High}\\ {\bf School}\\  {\bf Government}\\ {\bf \& Politics}} \\
\midrule
\multirow{6}{*}{Base} & 0 & 500K & 57.14 & 50.00 &  40.63 & 40.91 & 55.56 & 55.56 & 59.09 & 61.90 \\[0.2em]
& 250K & 500K & 57.14 & 50.00 & {\bf 50.00} & {\bf 50.00} & 55.56 & 55.56 & {\bf 68.18} & 61.90\\[0.2em]
& 120K & 500K & 57.14 & 50.00 & {\bf 50.00} & {\bf 50.00} & 55.56 & 55.56 & {\bf 63.64} & {\bf 66.67} \\[0.2em]
& 0 & 1M & 50.00 & 50.00 & 43.75 & 40.91 & 55.56 & 61.11 & 68.18 & 71.43\\[0.2em]
& 250K & 1M & {\bf 57.14} & 50.00 & 43.75 & {\bf 50.00} & 55.56 & 61.11 & {\bf 72.73} & 61.90\\[0.2em]
& 120K & 1M & {\bf 64.29} & {\bf 60.00} & {\bf 46.88} & {\bf 45.45} & 55.56 & 55.56 & {\bf 77.27} & 66.67\\
\midrule \midrule
\bf{Size} & \bf{$\tau$} & {\bf Ckpt} & \thead{{\bf High}\\ {\bf School}\\ {\bf Macro-}\\ {\bf economics}} & \thead{{\bf High}\\ {\bf School}\\ {\bf Mathe-}\\ {\bf matics}} & \thead{{\bf High}\\ {\bf School}\\ {\bf Micro-}\\ {\bf economics}} &  \thead{{\bf High}\\ {\bf School}\\ {\bf Physics}} & \thead{{\bf High}\\ {\bf School}\\ {\bf Psy-}\\ {\bf chology}}  & \thead{{\bf High}\\ {\bf School}\\ {\bf Statistics}} &  \thead{{\bf High}\\ {\bf School}\\ {\bf US}\\ {\bf History}}  &  \thead{{\bf High}\\ {\bf School}\\ {\bf World}\\ {\bf History}}\\
\midrule
\multirow{6}{*}{Base} & 0 & 500K & 39.53 & 41.38 & 46.15 &  41.18 &  45.00 & 47.83 & 68.18 & 53.85 \\[0.2em]
& 250K & 500K & 34.88 & 41.38 & {\bf 53.85} & {\bf 47.06} & {\bf 48.33} & 47.83 & 68.18 & 53.85\\[0.2em]
& 120K & 500K & 34.88 & 37.93 & {\bf 53.85} & {\bf 47.06} & {\bf 48.33} & 47.83 & 68.18 & {\bf 57.69} \\[0.2em]
& 0 & 1M & 37.21 & 41.38 & 53.85 & 41.18 & 50.00 & 47.83 & 72.73 & 57.69\\[0.2em]
& 250K & 1M & 37.21 & 41.38 & 53.85 & {\bf 47.06} & 46.67 & 47.83 & 68.18 & 53.85\\[0.2em]
& 120K & 1M & {\bf 41.86} & 41.38 & 53.85 & {\bf 47.06} & 50.00 & 43.48 & 72.73 & 57.69\\
\bottomrule
\end{tabular}
\end{center}
\end{adjustwidth}
\vspace{2mm}
\caption{Breakdown of first 32 of total 57 MMLU \citep{hendrycks2021measuring} tasks with direct answers for $\spactor_{\textrm{Base}}$. The metric are all accuracy.}
\label{tab:mmlu_breakdown_1}
\end{table}

\begin{table}[ht]
\begin{adjustwidth}{-.7in}{-.7in}
\begin{center}
\small
\setlength{\tabcolsep}{3.0pt}
\begin{tabular}{ccccccccccc}
\toprule
\bf{Size} & \bf{$\tau$} & {\bf Ckpt} & \thead{{\bf Human}\\ {\bf Aging}} &  \thead{{\bf Human}\\ {\bf Sexuality}} &  \thead{{\bf Inter-}\\ {\bf national}\\ {\bf Law}} &  \thead{{\bf Juris-}\\ {\bf prudence}} &  \thead{{\bf Logical}\\ {\bf Fallacies}} &  \thead{{\bf Machine}\\ {\bf Learning}} &  \thead{{\bf Management}} &  \thead{{\bf Marketing}} \\
\midrule
\multirow{6}{*}{Base} & 0 & 500K & 39.13 & 50.00 & 61.54 & 45.45 & 55.56 & 45.45 &   63.64 & 68.00\\[0.2em]
& 250K & 500K & 39.13 & 50.00 & {\bf 69.23} & 36.36 & {\bf 61.11} & 45.45 & 63.64 & 64.00\\[0.2em]
& 120K & 500K & {\bf 47.83} & {\bf 58.33} & 61.54 & 36.36 & {\bf 61.11} & 45.45 & 54.55 & 64.00 \\[0.2em]
& 0 & 1M & 43.48 & 58.33 & 69.23 & 45.45 & 66.67 & 45.45 & 63.64 & 68.00\\[0.2em]
& 250K & 1M & 43.48 & 50.00 & 61.54 & 45.45 & 61.11 & {\bf 63.64} & 63.64 & 68.00\\[0.2em]
& 120K & 1M & {\bf 47.83} & 50.00 & 61.54 & 45.45 & {\bf 72.22} & 45.45 & 63.64 & {\bf 72.00}\\
\midrule \midrule
\bf{Size} & \bf{$\tau$} & {\bf Ckpt} & \thead{{\bf Medical}\\ {\bf Genetics}} &  {\bf Misc.} &  \thead{{\bf Moral}\\ {\bf Disputes}} &  \thead{{\bf Moral}\\ {\bf Scenarios}} &  {\bf Nutrition} &  {\bf Philosophy} &  {\bf Prehistory} &  \thead{{\bf Professional}\\ {\bf Accounting}}\\
\midrule
\multirow{6}{*}{Base} & 0 & 500K & 45.45 & 39.53 & 50.00 & 33.00 &  51.52 & 35.29 & 48.57 & 35.48 \\[0.2em]
& 250K & 500K & {\bf 54.55} & 38.37 & 50.00 & 33.00 & 51.52 & {\bf 41.18} & 48.57 & 35.48\\[0.2em]
& 120K & 500K & {\bf 63.64} & {\bf 40.70} & 44.74 & 33.00 & 51.52 & {\bf 41.18} & 45.71 & 35.48\\[0.2em]
& 0 & 1M & 54.55 & 40.70 & 50.00 & 32.00 & 57.58 & 38.24 & 48.57 & 35.48\\[0.2em]
& 250K & 1M & {\bf 63.64} & 40.70 & 50.00 & {\bf 33.00} & 57.58 & {\bf 41.18} & {\bf 54.29} & {\bf 41.94}\\[0.2em]
& 120K & 1M & 54.55 & 39.53 & 47.37 & {\bf 34.00} & 54.55 & {\bf 41.18} & 45.71 & 35.48\\
\midrule \midrule
\bf{Size} & \bf{$\tau$} & {\bf Ckpt} &  \thead{{\bf Professional}\\ {\bf Law}} &  \thead{{\bf Professional}\\ {\bf Medicine}} &  \thead{{\bf Professional}\\ {\bf Psychology}} &  \thead{{\bf Public}\\ {\bf Relations}} &  \thead{{\bf Security}\\ {\bf Studies}} &  {\bf  Sociology} &  \thead{{\bf US Foreign}\\ {\bf Policy}} &  {\bf Virology}\\
\midrule
\multirow{6}{*}{Base} & 0 & 500K & 35.29 & 38.71 & 46.38 & 75.00 & 44.44 &  63.64 & 63.64 & 50.00 \\[0.2em]
& 250K & 500K & 35.29 & 38.71 & 44.93 & 58.33 & {\bf 48.15} & 63.64 & 54.55 & {\bf 55.56}\\[0.2em]
& 120K & 500K & 35.29 & 35.48 & {\bf 49.28} & 66.67 & 40.74 & 63.64 & 54.55 & {\bf 55.56} \\[0.2em]
& 0 & 1M & 35.29 & 38.71 & 44.93 & 66.67 & 48.15 & 68.18 & 63.64 & 50.00\\[0.2em]
& 250K & 1M & 33.53 & 38.71 & {\bf 49.28} & 58.33 & 44.44 & 68.18 & 63.64 & {\bf 61.11}\\[0.2em]
& 120K & 1M & 33.53 & 38.71 & {\bf 47.83} & 66.67 & 48.15 & 68.18 & 63.64 & {\bf 61.11} \\
\midrule \midrule
bf{Size} & \bf{$\tau$} & {\bf Ckpt} & \thead{{\bf World}\\ {\bf Religions}} & & & & & & & \\
\midrule
\multirow{6}{*}{Base} & 0 & 500K & 42.11 & & & & & & & \\[0.2em]
& 250K & 500K & 36.84 & & & & & & & \\[0.2em]
& 120K & 500K & {\bf 47.37} & & & & & & & \\[0.2em]
& 0 & 1M & 47.37 & & & & & & & \\[0.2em]
& 250K & 1M & 47.37 & & & & & & & \\[0.2em]
& 120K & 1M & 47.37 & & & & & & & \\
\bottomrule
\end{tabular}
\end{center}
\end{adjustwidth}
\vspace{2mm}
\caption{Breakdown of second 25 of total 57 MMLU \citep{hendrycks2021measuring} tasks with direct answers for $\spactor_{\textrm{Base}}$. The metric are all accuracy.}
\label{tab:mmlu_breakdown_2}
\end{table}

\begin{table}[ht]
\begin{adjustwidth}{-.7in}{-.7in}
\begin{center}
\small
\setlength{\tabcolsep}{3.0pt}
\begin{tabular}{ccccccccccc}
\toprule
\bf{Size} & \bf{$\tau$} & {\bf Ckpt} &  \thead{{\bf Abstract}\\ {\bf Algebra}} &  \thead{{\bf Anatomy}} &  \thead{\bf Astronomy} &  \thead{{\bf Business}\\ {\bf Ethics}} &  \thead{{\bf Clinical}\\ {\bf Knowledge}} &  \thead{{\bf College}\\ {\bf Biology}} &  \thead{{\bf College}\\ {\bf Chemistry}} &  \thead{{\bf College}\\ {\bf Computer}\\  {\bf Science}}  \\
\midrule
\multirow{4}{*}{Large} & 0 & 500K & 36.36 & 57.14 & 56.25 & 72.73 & 55.17 & 43.75 & 50.00 & 54.55 \\[0.2em]
& 120K & 500K & {\bf 45.45} & {\bf 57.14} & 50.00 & {\bf 72.73} & {\bf 62.07} & {\bf 62.50} & {\bf 50.00} & {\bf 63.64} \\[0.2em]
& 0 & 1M & 45.45 & 50.00 & 56.25 & 72.73 & 55.17 & 56.25 & 50.00 & 54.55 \\[0.2em]
& 120K & 1M & {\bf 54.55} & {\bf 64.29} & 43.75 & {\bf 72.73} & {\bf 58.62} & {\bf 62.50} & {\bf 50.00} & {\bf 54.55} \\
\midrule \midrule
\bf{Size} & \bf{$\tau$} & {\bf Ckpt} &  \thead{{\bf College}\\ {\bf Mathematics}} & \thead{{\bf College}\\ {\bf Medicine}} &  \thead{{\bf College}\\ {\bf Physics}} &  \thead{{\bf Computer}\\ {\bf Security}} &  \thead{{\bf Conceptual}\\ {\bf Physics}} &  \thead{{\bf Econo-}\\ {\bf metrics}} &  \thead{{\bf Electrical}\\ {\bf Engineering}} & \thead{{\bf  Elementary}\\ {\bf Mathematics}} \\
\midrule
\multirow{4}{*}{Large} & 0 & 500K & 45.45 & 59.09 & 81.82 & 36.36 & 42.31 & 50.00 & 62.50 & 39.02 \\[0.2em]
& 120K & 500K & {\bf 54.55} & {\bf 63.64} & 72.73 & {\bf 54.55} & {\bf 46.15} & {\bf 58.33} & {\bf 62.50} & 36.59 \\[0.2em]
& 0 & 1M & 54.55 & 59.09 & 81.82 & 54.55 & 42.31 & 50.00 & 68.75 & 39.02 \\[0.2em]
& 120K & 1M & 36.36 & 54.55 & {\bf 90.91} & 45.45 & {\bf 50.00} & {\bf 58.33} & 62.50 & {\bf 43.90} \\
\midrule \midrule
\bf{Size} & \bf{$\tau$} & {\bf Ckpt} & \thead{{\bf Formal}\\ {\bf Logic}} & \thead{{\bf Global}\\ {\bf Facts}} & \thead{{\bf High}\\ {\bf School}\\  {\bf Biology}} & \thead{{\bf High}\\ {\bf School}\\  {\bf Chemistry}} & \thead{{\bf High}\\ {\bf School}\\  {\bf Computer}\\ {\bf Science}} & \thead{{\bf High}\\ {\bf School}\\  {\bf European}\\ {\bf History}} & \thead{{\bf High}\\ {\bf School}\\  {\bf Geography}} & \thead{{\bf High}\\ {\bf School}\\  {\bf Government}\\ {\bf \& Politics}} \\
\midrule
\multirow{4}{*}{Large} & 0 & 500K & 57.14 & 60.00 & 46.88 & 45.45 & 55.56 & 72.22 & 77.27 & 61.90 \\[0.2em]
& 120K & 500K & 50.00 & {\bf 60.00} & {\bf 46.88} & {\bf 45.45} & {\bf 66.67} & {\bf 72.22} & {\bf 77.27} & {\bf 66.67} \\[0.2em]
& 0 & 1M & 64.29 & 60.00 & 46.88 & 45.45 & 66.67 & 66.67 & 81.82 & 71.43 \\[0.2em]
& 120K & 1M & 50.00 & {\bf 70.00} & 43.75 & 36.36 & {\bf 66.67} & {\bf 72.22} & {\bf 81.82} & 61.90 \\
\midrule \midrule
\bf{Size} & \bf{$\tau$} & {\bf Ckpt} & \thead{{\bf High}\\ {\bf School}\\ {\bf Macro-}\\ {\bf economics}} & \thead{{\bf High}\\ {\bf School}\\ {\bf Mathe-}\\ {\bf matics}} & \thead{{\bf High}\\ {\bf School}\\ {\bf Micro-}\\ {\bf economics}} &  \thead{{\bf High}\\ {\bf School}\\ {\bf Physics}} & \thead{{\bf High}\\ {\bf School}\\ {\bf Psy-}\\ {\bf chology}}  & \thead{{\bf High}\\ {\bf School}\\ {\bf Statistics}} &  \thead{{\bf High}\\ {\bf School}\\ {\bf US}\\ {\bf History}}  &  \thead{{\bf High}\\ {\bf School}\\ {\bf World}\\ {\bf History}}\\
\midrule
\multirow{4}{*}{Large} & 0 & 500K &  41.86 & 48.28 & 65.38 & 47.06 & 63.33 & 47.83 & 59.09 & 57.69 \\[0.2em]
& 120K & 500K & {\bf 44.19} & {\bf 48.28} & 57.69 & {\bf 47.06} & {\bf 66.67} & 43.48 & {\bf 68.18} & {\bf 69.23} \\[0.2em]
& 0 & 1M & 44.19 & 44.83 & 65.38 & 47.06 & 66.67 & 52.17 & 59.09 & 65.38 \\[0.2em]
& 120K & 1M & 41.86 & {\bf 44.83} & {\bf 69.23} & {\bf 52.94} & {\bf 70.00} & {\bf 56.52} & {\bf 68.18} & 61.54 \\
\bottomrule
\end{tabular}
\end{center}
\end{adjustwidth}
\vspace{2mm}
\caption{Breakdown of first 32 of total 57 MMLU \citep{hendrycks2021measuring} tasks with direct answers for $\spactor_{\textrm{Large}}$. The metric are all accuracy.}
\label{tab:mmlu_breakdown_large_1}
\end{table}

\begin{table}[ht]
\begin{adjustwidth}{-.7in}{-.7in}
\begin{center}
\small
\setlength{\tabcolsep}{3.0pt}
\begin{tabular}{ccccccccccc}
\toprule
\bf{Size} & \bf{$\tau$} & {\bf Ckpt} & \thead{{\bf Human}\\ {\bf Aging}} &  \thead{{\bf Human}\\ {\bf Sexuality}} &  \thead{{\bf Inter-}\\ {\bf national}\\ {\bf Law}} &  \thead{{\bf Juris-}\\ {\bf prudence}} &  \thead{{\bf Logical}\\ {\bf Fallacies}} &  \thead{{\bf Machine}\\ {\bf Learning}} &  \thead{{\bf Management}} &  \thead{{\bf Marketing}} \\
\midrule
\multirow{4}{*}{Large} & 0 & 500K & 47.83 & 58.33 & 76.92 & 54.55 & 66.67 & 45.45 & 72.73 & 84.00 \\[0.2em]
& 120K & 500K & {\bf 52.17} & {\bf 66.67} & {\bf 76.92} & {\bf 63.64} & {\bf 77.78} & 36.36 & {\bf 72.73} & {\bf 84.00} \\[0.2em]
& 0 & 1M & 56.52 & 66.67 & 84.62 & 54.55 & 72.22 & 36.36 & 81.82 & 84.00 \\[0.2em]
& 120K & 1M & {\bf 56.52} & {\bf 66.67} & 76.92 & {\bf 63.64} & {\bf 72.22} & {\bf 36.36} & 72.73 & {\bf 88.00} \\
\midrule \midrule
\bf{Size} & \bf{$\tau$} & {\bf Ckpt} & \thead{{\bf Medical}\\ {\bf Genetics}} &  {\bf Misc.} &  \thead{{\bf Moral}\\ {\bf Disputes}} &  \thead{{\bf Moral}\\ {\bf Scenarios}} &  {\bf Nutrition} &  {\bf Philosophy} &  {\bf Prehistory} &  \thead{{\bf Professional}\\ {\bf Accounting}}\\
\midrule
\multirow{4}{*}{Large} & 0 & 500K & 63.64 & 51.16 & 50.00 & 32.00 & 57.58 & 47.06 & 48.57 & 41.94 \\[0.2em]
& 120K & 500K & {\bf 72.73} & 50.00 & {\bf 55.26} & {\bf 32.00} & 54.55 & {\bf 52.94} & {\bf 57.14} & 38.71 \\[0.2em]
& 0 & 1M & 72.73 & 52.33 & 52.63 & 40.00 & 51.52 & 50.00 & 57.14 & 51.61 \\[0.2em]
& 120K & 1M & 63.64 & {\bf 52.33} & 50.00 & 34.00 & {\bf 54.55} & {\bf 52.94} & {\bf 62.86} & 45.16 \\
\midrule \midrule
\bf{Size} & \bf{$\tau$} & {\bf Ckpt} &  \thead{{\bf Professional}\\ {\bf Law}} &  \thead{{\bf Professional}\\ {\bf Medicine}} &  \thead{{\bf Professional}\\ {\bf Psychology}} &  \thead{{\bf Public}\\ {\bf Relations}} &  \thead{{\bf Security}\\ {\bf Studies}} &  {\bf  Sociology} &  \thead{{\bf US Foreign}\\ {\bf Policy}} &  {\bf Virology}\\
\midrule
\multirow{4}{*}{Large} & 0 & 500K & 33.53 & 48.39 & 56.52 & 66.67 & 48.15 & 72.73 & 63.64 & 61.11 \\[0.2em]
& 120K & 500K & {\bf 34.12} & {\bf 51.61} & 55.07 & {\bf 66.67} & {\bf 51.85} & 63.64 & 54.55 & {\bf 72.22} \\[0.2em]
& 0 & 1M & 32.94 & 58.06 & 57.97 & 75.00 & 44.44 & 77.27 & 81.82 & 61.11 \\[0.2em]
& 120K & 1M & {\bf 34.12} & 54.84 & {\bf 62.32} & {\bf 75.00} & {\bf 55.56} & {\bf 77.27} & 72.73 & {\bf 66.67}\\
\midrule \midrule
bf{Size} & \bf{$\tau$} & {\bf Ckpt} & \thead{{\bf World}\\ {\bf Religions}} & & & & & & & \\
\midrule
\multirow{4}{*}{Large} & 0 & 500K & 57.89 \\[0.2em]
& 120K & 500K & {\bf 57.89} \\[0.2em]
& 0 & 1M & 63.16 \\[0.2em]
& 120K & 1M & {\bf 68.42}\\
\bottomrule
\end{tabular}
\end{center}
\end{adjustwidth}
\vspace{2mm}
\caption{Breakdown of second 25 of total 57 MMLU \citep{hendrycks2021measuring} tasks with direct answers for $\spactor_{\textrm{Large}}$. The metric are all accuracy.}
\label{tab:mmlu_breakdown_large_2}
\end{table}

\end{document}